\documentclass[runningheads]{llncs}

\usepackage{eccv}

\usepackage{eccvabbrv}

\usepackage{graphicx}
\usepackage{booktabs}
\usepackage{sidecap}

\usepackage[table]{xcolor}
\usepackage{multicol}
\usepackage{multirow}

\usepackage[accsupp]{axessibility}  

\usepackage{hyperref}

\usepackage{orcidlink}

\newcommand\blfootnote[1]{%
  \begingroup
  \renewcommand\thefootnote{}\footnotetext{#1}%
  \addtocounter{footnote}{-1}%
  \endgroup
}

\begin{document}

\title{OCA: ODE-Driven Cross-Attention for Image-to-\\Point-Cloud Registration} 

\titlerunning{OCA: ODE-Driven Cross-Attention for Image-to-Point-Cloud Registration}

\author{Pei An\inst{1}\orcidlink{0000-0002-3645-8465} \and
Jiaqi Yang \inst{2}\orcidlink{0000-0002-2071-2457}
\and
Yulong Wang \inst{3}\orcidlink{0000-0002-1033-9281}
\and
Siwen Quan \inst{4\dag}\orcidlink{0000-0001-7579-937X}
\and\\
Liangliang Nan \inst{5\dag}\orcidlink{0000-0002-5629-9975}
}

\authorrunning{P.~An et al.}

\institute{Huazhong University of Science and Technology, China
\email{anpei96@hust.edu.cn} 
\and
Northwestern Polytechnical University, China
\email{jqyang@nwpu.edu.cn}
\and
Huazhong Agricultural University, China
\email{ylwang@mail.hzau.edu.cn}
\and
Chang’an University, China
\email{siwenquan@chd.edu.cn}
\and
Delft University of Technology, Netherlands
\email{liangliang.nan@tudelft.nl}
}

\maketitle

\blfootnote{
$\dag$ Corresponding author(s).
}

\begin{abstract}
Cross-attention is a crucial component in learning-based image-to-point-cloud (I2P) registration. Although existing cross-attention mechanisms have achieved promising progress, attention ambiguity remains a fundamental challenge that hinders the learning of discriminative 2D-3D correspondences. To address this problem, we revisit cross-attention and establish ordinary differential equations (ODEs) to model the ideal I2P feature interaction. Based on this formulation, we develop an ODE-driven cross-attention (OCA) module that refines feature representations and attention matrices through ODEs. In practice, OCA can be seamlessly integrated into existing I2P registration frameworks. To validate its effectiveness, we incorporate OCA into five state-of-the-art baselines and evaluate on four public benchmark datasets. Experimental results demonstrate that OCA improves registration recall by up to 5\%, 9\%, and 15\% under the standard, fine-tuning, and zero-shot settings, respectively. Code is released at \url{github.com/anpei96/oca-i2p-demo}.

\end{abstract}

\section{Introduction}
\label{sec:introduction}


Image-to-point-cloud (I2P) registration is a fundamental task in computer vision \cite{p2-net}. Given an image and a 3D point cloud pair, it aims to establish 2D-3D pixel-to-point correspondences and estimate the camera pose within the point cloud coordinate system \cite{go-match-22}. Accordingly, I2P registration supports a wide range of applications, including visual localization \cite{vis-loc-cvpr-25, new_add_1}, state estimation \cite{state_est}, point cloud colorization \cite{colorization}, and simultaneous localization and mapping (SLAM) \cite{mast3r-slam}. Recently, learning-based I2P registration methods have attracted increasing attention \cite{2d3d-match, 2d3d-matr, corr-tcsvt}. In existing registration pipelines, feature interaction serves as a critical module. It bridges the modality gap and yields discriminative, modality-invariant features for robust I2P registration \cite{2d3d-matr, graph-i2p-cvpr, diff_reg_match, diff-i2p-iccv-25}.

\begin{figure}[t]
	\centering
		\includegraphics[width=1.0\linewidth]{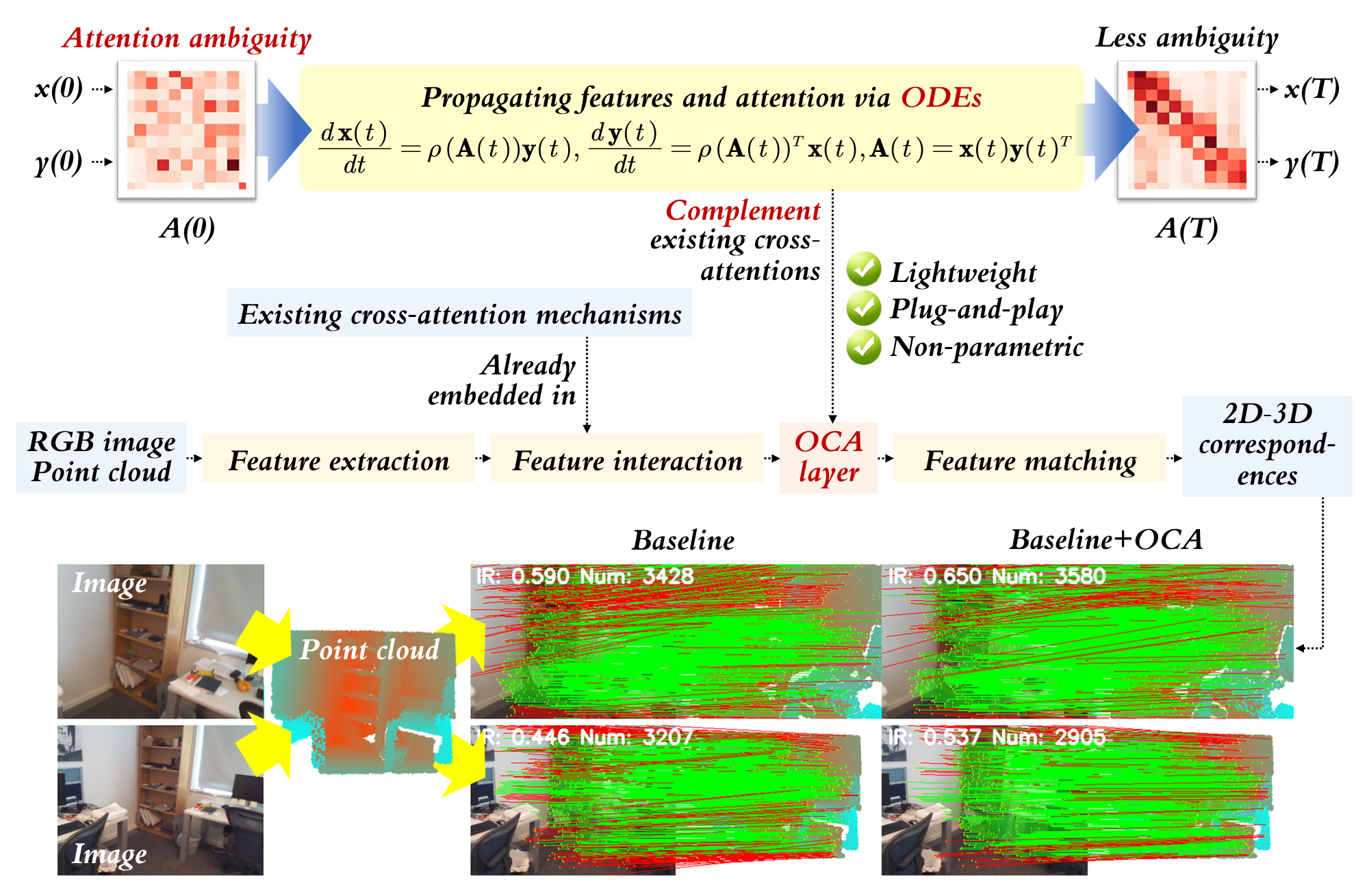}
		\caption{\textbf{Motivation} of ODE-driven cross-attention (OCA). Although existing cross-attention models have made progress, attention ambiguity caused by the modality gap remains inevitable. To overcome this problem, we propose OCA to enhance I2P feature representations by propagating features and the attention matrix through ODEs. IR denotes the inlier ratio. \textcolor{green}{\textbf{Green}} and \textcolor{red}{\textbf{red}} lines represent inliers and outliers, respectively.}
	\label{fig:intro}
\end{figure}



Cross-attention serves as the core component in feature interaction. In early works, cross-attention was adopted from representative intra-modal registration frameworks, including SuperGlue \cite{superglue} and GeoTransformer \cite{geo-tf}. In these methods \cite{superglue,geo-tf}, cross-attention is employed to model the similarity of pairwise features between the source and target frames. Recently, cross-attention has been applied to I2P registration \cite{2d3d-matr} since it models the feature similarity between 2D pixels and 3D points, thereby facilitating discriminative and modality-invariant correspondence learning \cite{bridge}.



However, when applying cross-attention from intra-modal to cross-modal I2P registration, \textbf{attention ambiguity} becomes an inevitable challenge. Due to the large modality gap, spurious 2D-3D correspondences may exhibit high similarity scores, which impedes the learning of reliable cross-modal correspondences \cite{bridge, flowi2p}. To mitigate such ambiguity, researchers have improved cross-attention from various aspects, such as feature manifold alignment \cite{flowi2p}, feature uncertainty correction \cite{bridge}, feature covariance alignment \cite{ca-i2p-iccv-25}, and keypoint-based correspondence learning \cite{ldf-i2p}. Nevertheless, their capabilities (robustness and generalization ability) still have substantial room for improvement. 




To strengthen the existing cross-attention mechanisms \cite{2d3d-matr, flowi2p, bridge, ca-i2p-iccv-25, ldf-i2p}, we propose a plug-and-play module called \textbf{ODE-driven cross-attention} (OCA) (Fig. \ref{fig:intro}). To alleviate attention ambiguity, we establish the ordinary differential equations (ODEs) termed assignment ODEs to mimic the ideal feature interaction. By analyzing the convergence conditions of the assignment ODEs, we develop the OCA module to refine I2P 
feature representations by propagating features and the attention matrix through ODEs. As a lightweight and non-parametric module, OCA can complement existing cross-attentions by integrating itself into current I2P registration frameworks. To validate the effectiveness of OCA, we perform extensive experiments on four standard datasets, namely 7-Scenes \cite{scene-7-dataset}, RGBD-v2 \cite{RGBD-dataset}, TUM \cite{tum}, and ScanNet \cite{scan-net}, across five state-of-the-art baselines with advanced cross-attention schemes \cite{2d3d-matr, flowi2p, bridge, ca-i2p-iccv-25, ldf-i2p}. Results demonstrate that OCA improves the registration recall by up to \textbf{5\%, 9\%, and 15\%} under the standard, fine-tuning, and zero-shot evaluation settings. In summary, our main contributions are as follows:


\begin{itemize}
\item To approximate the ideal feature interaction, we construct the assignment ODEs, enabling the analysis of cross-attention from an ODE perspective.
\item By analyzing the convergence condition of assignment ODEs, we propose a lightweight ODE-driven cross-attention (OCA) module to iteratively refine I2P representations through ODE propagation.
\item OCA can be seamlessly
integrated into existing I2P registration frameworks and complement the current cross-attention modules.
\end{itemize}

\section{Related works}
\label{sec:related_work}

We provide a brief review of learning-based I2P registration and discuss the development of cross-attention for I2P registration. 

\vspace{+1mm}
\noindent\textbf{Learning based I2P registration}. To mitigate the modal discrepancy between images and point clouds, deep learning has become the dominant paradigm for image-to-point (I2P) registration. In 2019, Feng et al. \cite{2d3d-match} introduced the first learning-based I2P registration framework. Inspired by work \cite{2d3d-match}, P2-Net \cite{p2-net}, and Deep-I2P \cite{deep-i2p} were subsequently developed, which learn modality-invariant features using separate encoders for each modality \cite{resnet, kpconv}. To suppress 2D-3D outliers, circle loss \cite{circle-loss} is used to learn discriminative 2D-3D descriptors \cite{p2-net}. Since 2023, researchers have identified the lack of effective I2P feature interaction as the key bottleneck in suboptimal cross-modal representation learning \cite{corr-tcsvt, go-match-22, 2d3d-matr, diff_reg_match, diff-reg}. Accordingly, transformer-based cross-attention has been integrated into existing I2P registration architectures \cite{2d3d-matr}, as illustrated in the middle image of Fig. \ref{fig:intro}. By designing dedicated feature interaction modules, recent I2P registration works \cite{flowi2p, bridge, ca-i2p-iccv-25} have achieved substantial performance gains on public benchmarks.

\vspace{+1mm}
\noindent\textbf{Cross-attention in I2P registration}. Above works \cite{2d3d-matr, flowi2p, bridge, ca-i2p-iccv-25} have firmly established that cross-attention is indispensable for high-performance I2P registration. In this section, we review the evolution of cross-attention. Early approaches directly apply cross-attention between 2D and 3D keypoints \cite{go-match-22, corr-tcsvt, corr-learning-i2p-cvpr}. However, without reliable priors, it is challenging to extract robust 2D and 3D keypoints that retain sufficient true inliers. To avoid this issue, a naive alternative is to perform cross-attention between all 2D pixels and all 3D points, but this incurs an excessively heavy computational burden \cite{go-match-22}. A practical compromise is to conduct cross-attention on 2D and 3D patches \cite{2d3d-matr}. This solution has been widely adopted in state-of-the-art I2P registration frameworks \cite{diff_reg_match, flowi2p, bridge, diff-i2p-iccv-25, ca-i2p-iccv-25}. 

To pursue higher performance, researchers have recently developed advanced cross-attention to address the attention ambiguity. Current methods can be divided into two categories: prior-knowledge-based and prior-knowledge-free. Prior-knowledge-based cross-attention is to leverage geometric and semantic priors to suppress noisy feature correlations during cross-attention \cite{diff-reg, top-i2p, diff-i2p-iccv-25}. However, such methods are sensitive to the quality and generalization of pre-trained visual foundation models. Prior-knowledge-free based cross-attention is to design the more expressive deep neural networks \cite{2d3d-matr, flowi2p, bridge, ca-i2p-iccv-25, ldf-i2p}, as discussed in Sec. \ref{sec:introduction}. These methods do not rely on external pre-trained models, making them more convenient for real-world deployment.  

In this paper, we attempt to strengthen the prior-knowledge-free based cross-attention from an ODE perspective. Experimental results in Sec. \ref{sec:exp} verify that our approach outperforms representative prior-free methods \cite{2d3d-matr, flowi2p, bridge, ca-i2p-iccv-25, ldf-i2p}.

\section{Rethinking cross-attention from ODEs}
\label{sec:explain}

\begin{figure}[t]
	\centering
		\includegraphics[width=1.0\linewidth]{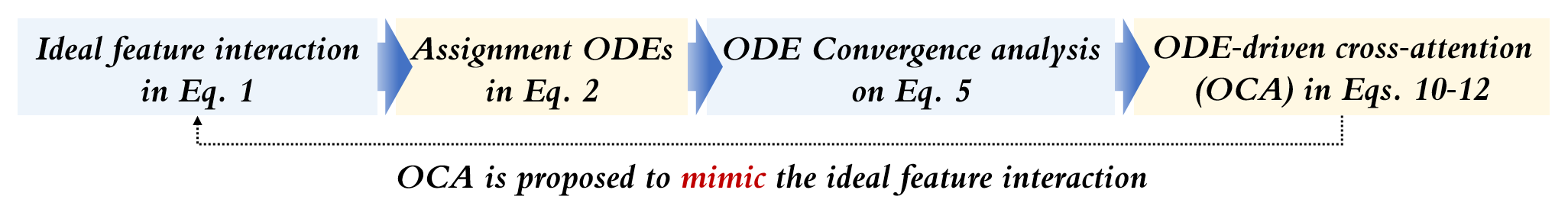}
		\caption{Overview of Sec. \ref{sec:explain} and \ref{sec:refine}. OCA is designed to alleviate attention ambiguity by approximating the ideal feature interaction formulated in Eq. \ref{eq:assign_eq_ideal}.}
	\label{fig:framework}
\end{figure}

Starting from the ideal feature interaction, we reformulate cross-attention from an ODE perspective (see Fig. \ref{fig:framework}). 

\subsection{Assignment ODEs}
\label{sec:explain_assign_eq}

To eliminate the modality difference of I2P data, we first model the procedure of {ideal feature interaction}. Let $\mathbf{x}\in \mathbb{R}^{N\times c}$ and $\mathbf{y}\in \mathbb{R}^{M\times c}$ be features of 2D pixels and 3D points, extracted separately from RGB image $\mathbf{I}\in\mathbb{R}^{H\times W\times 3}$ and point cloud $\mathbf{P} \in \mathbb{R}^{M\times 3}$. $c$ is the number of feature channels. $N=HW$ and $M$ are the numbers of pixels and points, respectively.  An {ideal feature interaction} can be constructed as: 
\begin{equation}
\label{eq:assign_eq_ideal}
\mathbf{x}'= \mathbf{x} + \mathbf{A}_{gt} \mathbf{y}, \,\,
\mathbf{y}'= \mathbf{y} + \mathbf{A}^T_{gt} \mathbf{x}
\end{equation}
where $\mathbf{A}_{gt} \in \{0,1\}^{N\times M}$ is the ground truth (GT) assignment matrix. $\mathbf{A}_{gt} \mathbf{1} \leq N$, $ \mathbf{A}_{gt}^T \mathbf{1} \leq M$. $[\mathbf{A}_{gt}]_{ij}=1$ indicates that the $i$-th pixel and $j$-th point is a valid I2P correspondence (marked as $\langle i,j \rangle\in\mathcal{C}$ and $\mathcal{C}$ is a set of all valid correspondences). Eq. \ref{eq:assign_eq_ideal} is the ideal interaction, because it ensures that $[\mathbf{x}']_i = [\mathbf{y}']_j$ for all $\langle i,j \rangle\in\mathcal{C}$, resulting in modality-invariant features after interaction. Since $\mathbf{A}_{gt}$ is unknown before registration, Eq. \ref{eq:assign_eq_ideal} cannot be directly used in practice. 

To mimic Eq. \ref{eq:assign_eq_ideal} in the correspondence learning, we model $\mathbf{x}'$, $\mathbf{y}'$, and $\mathbf{A}_{gt}$ as time-varying variables (i.e., $\mathbf{x}(t)$, $\mathbf{y}(t)$, and $\mathbf{A}(t)$) and develop a ODE-based feature interaction scheme:
\begin{equation}
\label{eq:assign_eq}
\frac{d \mathbf{x}(t)}{d t}= \rho(\mathbf{A}(t)) \mathbf{y}(t), \,\,
\frac{d \mathbf{y}(t)}{d t}= \rho(\mathbf{A}(t))^T \mathbf{x}(t), \,\,\mathbf{A}(t) = \mathbf{x}(t)\mathbf{y}(t)^T \in \mathbb{R}^{N\times M}
\end{equation}
where $\mathbf{A}(t)$ is a feature correlation matrix, which is widely used in a transformer. $\rho(\cdot)$ is a row-normalized operator for a matrix, i.e., softmax. $\rho(\mathbf{A}(t))$ describes the similarity of the $i$-th pixel and $j$-th point, serving as an approximation of $\mathbf{A}_{gt}$. Due to this reason, we name Eq. \ref{eq:assign_eq} as the assignment ODEs.

\vspace{+1mm}
\noindent\textbf{Discussions of assignment ODEs}: The motivation of introducing assignment ODEs is to address attention ambiguity from the ODE viewpoint. Given an initial attention $\mathbf{A}(0)$, {our goal is to reduce attention ambiguity via $\rho(\mathbf{A}(+\infty)) \approx\mathbf{A}_{gt}$}. To reach such a $\mathbf{A}(+\infty)$, we derive ODEs to describe the dynamics of $\mathbf{A}(t)$ and develop the OCA module by analyzing the convergence conditions of ODEs. More specifically, according to Eq. \ref{eq:assign_eq}, the interacted features are $\mathbf{x}(T)=\mathbf{x}(0)+\int_0^T \rho(\mathbf{A}(t)) \mathbf{y}(t) dt$ and $\mathbf{y}(T)=\mathbf{y}(0)+\int_0^T \rho(\mathbf{A}(t))^T \mathbf{x}(t) dt$. If the initial attention $\rho(\mathbf{A}(0))$ is closed to $\mathbf{A}_{gt}$, $\int_0^T \rho(\mathbf{A}(t)) \mathbf{y}(t) dt$ and $\int_0^T \rho(\mathbf{A}(t))^T \mathbf{x}(t)$ would {enhance the feature similarity} of correct correspondences inside $\mathbf{x}(T)$ and $\mathbf{y}(T)$. Then, $\rho(\mathbf{A}(t))$ tends to be sparser and approaches $\mathbf{A}_{gt}$. It indicates that {attention ambiguity can be alleviated via the propagation of assignment ODEs}.


\subsection{Relation of cross-attention and assignment ODEs}
\label{sec:cross_assign_eq}

In this subsection, we reveal the relation between cross-attention and assignment ODEs in Eq. \ref{eq:assign_eq}. Using the explicit Euler method, Eq. \ref{eq:assign_eq} can be reformulated in a discrete form under the approximation $\rho(\mathbf{A}(t))^T\approx \rho(\mathbf{A}(t)^T)$:
\begin{equation}
\label{eq:assign_eq_discrete}
\mathbf{x}[k+1] = \mathbf{x}[k]+ \rho(\mathbf{x}[k]\mathbf{y}[k]^T)\mathbf{y}[k], \,\,
\mathbf{y}[k+1] = \mathbf{y}[k]+ \rho(\mathbf{y}[k]\mathbf{x}[k]^T)\mathbf{x}[k]
\end{equation}

If $\rho(\cdot)$ is a softmax with a scaling factor (reducing matrix elements by $1/\sqrt{c}$ times), Eq. \ref{eq:assign_eq_discrete} can be formulated as a standard transformer-based cross-attention:
\begin{equation}
\label{eq:cross_att_layer}
\begin{aligned}
\mathbf{x}[k+1] &= \mathbf{x}[k]+ \mathtt{softmax}\left({\mathbf{x}[k]\mathbf{y}[k]^T}/{\sqrt{c}}\right)\mathbf{y}[k], \\
\mathbf{y}[k+1] &= \mathbf{y}[k]+ \mathtt{softmax}\left({\mathbf{y}[k]\mathbf{x}[k]^T}/{\sqrt{c}}\right)\mathbf{x}[k]
\end{aligned}
\end{equation}

The approximate equivalence between Eqs. \ref{eq:assign_eq_discrete} and \ref{eq:cross_att_layer} highlights the close connection between cross-attention and assignment ODEs. Consequently, analyzing the assignment ODEs can help overcome the attention ambiguity inherent in cross-attention.

\subsection{Convergence analysis of assignment ODEs}
\label{sec:convergence_assign_eq}

As discussed in Sec. \ref{sec:explain_assign_eq}, eliminating attention ambiguity is equivalent to determining whether $\rho(\mathbf{A}(t))$ converges to $\mathbf{A}_{gt}$ as $t$ is $+\infty$. Consequently, we analyze the convergence of $\rho(\mathbf{A}(t))$:
\begin{equation}
\label{eq:diff_assign_ode}
\frac{d\rho(\mathbf{A}(t))}{dt} = \frac{d\rho(\mathbf{A}(t))}{\mathbf{A}(t)} \frac{d\mathbf{A}(t)}{dt} = \rho'(\mathbf{A}(t))\left(\rho(\mathbf{A}(t)) \mathbf{Y}(t) + \mathbf{X}(t) \rho(\mathbf{A}(t)) \right)
\end{equation}
\begin{equation}
\label{eq:diff_assign_ode_aux}
\mathbf{X}(t)=\mathbf{x}(t)\mathbf{x}(t)^T,\,\,\mathbf{Y}(t)=\mathbf{y}(t)\mathbf{y}(t)^T
\end{equation}
$\rho'(\mathbf{A}(t))$ is the abbreviation of $d\rho(\mathbf{A}(t))/d\mathbf{A}(t)$. The derivation of Eq. \ref{eq:diff_assign_ode} is shown in Appendix A. To ensure the convergence of $\rho(\mathbf{A}(t))$ (i.e., $d\rho(\mathbf{A}(t))/dt=\mathbf{0}$), a stationary condition can arise when one of the following holds:

\vspace{+1mm}
\noindent\textbf{Condition (C1): $\rho'(\mathbf{A}(t))=\mathbf{0}$}. Since $\rho(\cdot)$ is a normalization operator, $\rho'(\mathbf{A}(t))=\mathbf{0}$ if $\mathbf{A}(t)$ is a permutation matrix. 

\noindent\textbf{Condition (C2): $\rho(\mathbf{A}(t)) \mathbf{Y}(t) + \mathbf{X}(t) \rho(\mathbf{A}(t))=\mathbf{0}$}. According to the theory of the Sylvester equation \cite{Sylvester}, non-zero $\rho(\mathbf{A}(t))$ can exist if $\mathbf{X}(t)$ and $\mathbf{Y}(t)$ have at least one same eigenvalue. In practice, this condition does not contribute to improving I2P registration performance (see discussion in Appendix A). 

\vspace{+1mm}
Motivated by this analysis, we derive, from \textbf{condition (C1)}, two practical strategies that encourage $\rho(\mathbf{A}(t))$ to approximate $\mathbf{A}_{gt}$:  

\vspace{+1mm}
\noindent\textbf{Strategy (S1)}: Ensure the sparsity of $\rho(\mathbf{A}(t))$, since a sparse $\rho(\mathbf{A}(t))$ can bring $\rho'(\mathbf{A}(t))$ close to a zero matrix. 

\noindent\textbf{Strategy (S2)}: Ensure that the initial value $\rho(\mathbf{A}(0))$ is close to $\mathbf{A}_{gt}$.  

\vspace{+1mm}
The standard transformer-based cross-attention defined in Eq.~\ref{eq:cross_att_layer} struggles to satisfy strategies \textbf{(S1)} and \textbf{(S2)}. The reasons are twofold. First, the softmax function cannot guarantee the sparsity of $\rho(\mathbf{A}(t))$. Second, it is difficult to ensure that $\rho(\mathbf{A}(0))$ is close to $\mathbf{A}_{gt}$, as feature matching relies on $L2$-normalized features. Thus, the standard cross-attention fails to effectively resolve attention ambiguity. 


\section{Refining cross-attention with ODEs}
\label{sec:refine}

From the analysis of assignment ODEs, we design the ODE-driven cross-attention (OCA) module and develop an OCA-embedded I2P registration framework. 

\begin{figure}[t]
	\centering
		\includegraphics[width=1.0\linewidth]{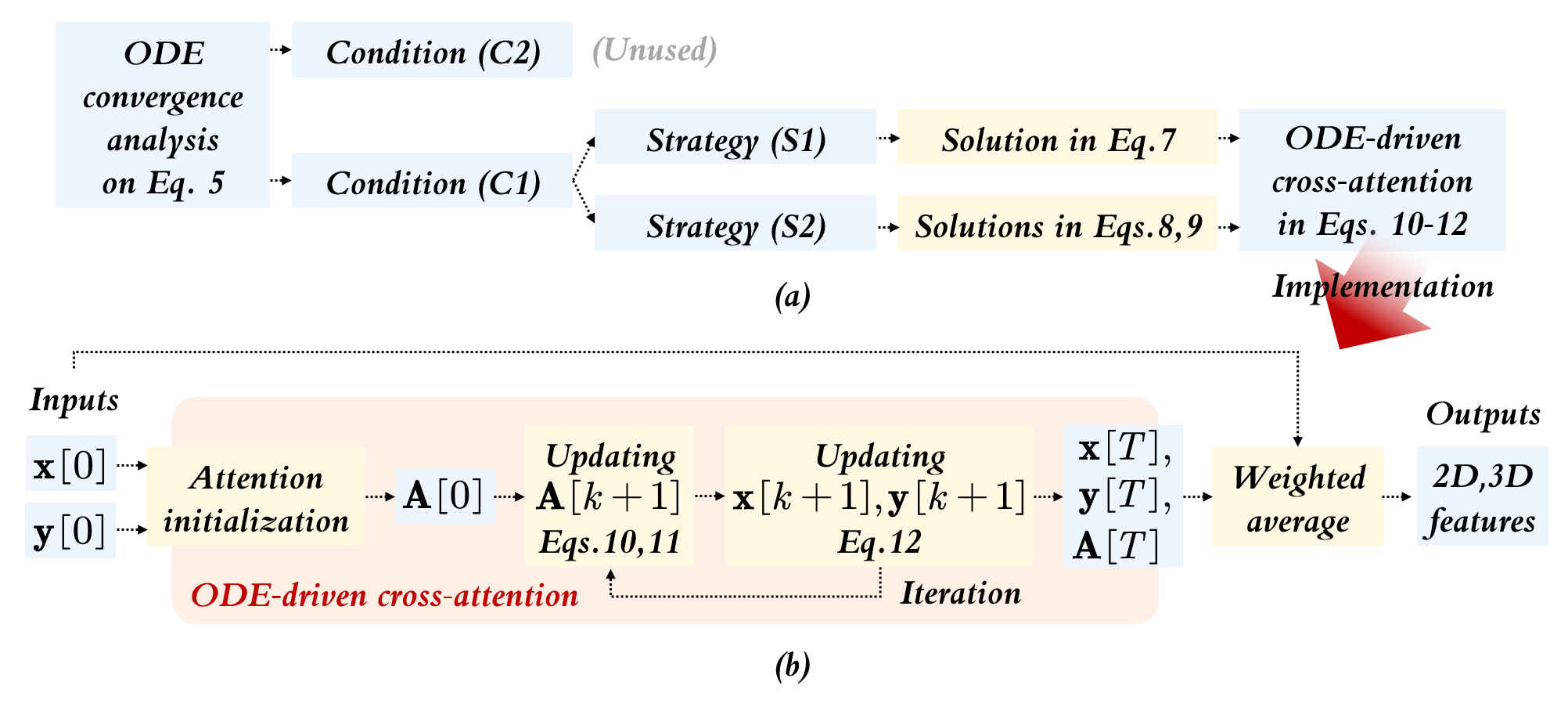}
		\caption{Overview of OCA. (a) From the convergence condition (C1), we discover that solutions in Eqs. \ref{eq:refine_prune}, \ref{eq:mat_a_new}, and \ref{eq:ref_propagate_a_mat} can guide the assignment ODEs to rapidly approach the ground-truth assignment matrix. After that, we propose OCA to unify the solutions in Eqs. \ref{eq:refine_prune}, \ref{eq:mat_a_new}, and \ref{eq:ref_propagate_a_mat}. (b) The overall pipeline of OCA. For computation efficiency, Eqs. \ref{eq:diff_assign_ode_aux_1}, \ref{eq:att_prop_1}, and \ref{eq:att_prop_2} are adopted to approximate Eqs. \ref{eq:mat_a_new} and \ref{eq:ref_propagate_a_mat}.}
	\label{fig:solution}
\end{figure}

\subsection{Feasible solutions from ODE perspective}
\label{sec:refine_overview}

To improve cross-attention from strategies \textbf{(S1)} and \textbf{(S2)}, we derive the feasible solutions that form the foundation of OCA. Towards \textbf{(S1)}, a simple strategy is to prune $\rho(\mathbf{A}(t))$. One naive way is the {top-K} based pruning. In the $i$-th row of $\rho(\mathbf{A}(t))$, the largest top-K elements are retained while others are set to zero. This procedure is described as:
\begin{equation}
\label{eq:refine_prune}
\rho_{\text{sparse}}(\mathbf{A}(t)) = \mathtt{norm\_row}(\mathtt{prune}(\mathtt{softmax}(\mathbf{A}(t)), K)) \approx \mathtt{softmax}(\gamma \mathbf{A}(t))
\end{equation}
where $\mathtt{norm\_row}(\cdot)$ is a function that normalizes each row of a matrix by dividing the sum of the elements in this row. Nevertheless, this scheme risks gradient vanishing in the initial training phase, since the correct correspondence easily does not belong to the top-K elements. A safe pruning scheme incorporates the temperature parameter $\gamma$ into the softmax function (right-hand side in Eq. \ref{eq:refine_prune}). $\gamma\geq1$ amplifies the difference of each row in $\mathbf{A}(t)$ so that $\mathtt{softmax}(\gamma \mathbf{A}(t))$ can be regarded as an approximation of $\rho_{\text{sparse}}(\mathbf{A}(t))$. Towards \textbf{(S2)}, we redefine the computation of $\mathbf{A}(t)$ as:
\begin{equation}
\label{eq:mat_a_new}
\mathbf{A}(t) = \mathbf{x}_{\text{norm}}(t)\mathbf{y}_{\text{norm}}(t)^T = \Vert \mathbf{x}(t)\Vert_2^{-1}\mathbf{x}(t) \Vert \mathbf{y}(t)\Vert_2^{-1}\mathbf{y}(t)^T
\end{equation}
The reason why Eq. \ref{eq:mat_a_new} is more accurate than Eq. \ref{eq:assign_eq} is that 2D-3D correspondence is identified based on the L2-normalized feature distance \cite{p2-net,2d3d-matr}, whereas standard transformers do not normalize features during cross-attention computation. 

Beyond Eqs. \ref{eq:refine_prune} and \ref{eq:mat_a_new}, the most crucial solution is to leverage ODEs in Eqs. \ref{eq:assign_eq} and \ref{eq:diff_assign_ode}. Given the initial values $\mathbf{x}(0)$, $\mathbf{y}(0)$, and $\mathbf{A}(0)$, we can forecast $\mathbf{x}(T)$, $\mathbf{y}(T)$, and $\mathbf{A}(T)$ by integrating $d\mathbf{x}(t)/dt$, $d\mathbf{y}(t)/dt$, and $d\mathbf{A}(t)/dt$. During the integrations, the feature similarity of incorrect correspondences is progressively suppressed\footnote{The reasons are two-fold. First, Eqs. \ref{eq:refine_prune} and \ref{eq:mat_a_new} ensure an initial attention matrix with the lower ambiguity. Second, the integration process suppresses the outliers (see the discussion of assignment ODEs in Sec. \ref{sec:explain_assign_eq}). Hence, propagating ODEs in Eqs. \ref{eq:assign_eq} and \ref{eq:diff_assign_ode} is helpful to alleviating the attention ambiguity.}.  In the above integrations, $\int d\mathbf{A}(t)/dt$ should be carefully addressed:
\begin{equation}
\label{eq:ref_propagate_a_mat}
\mathbf{A}(T) = \mathbf{A}(0)+\int_{0}^T d\mathbf{A}(t) = \mathbf{A}(0)+\int_{0}^T \rho(\mathbf{A}(t)) \mathbf{Y}(t) + \mathbf{X}(t) \rho(\mathbf{A}(t)) dt
\end{equation}
If we use Eq. \ref{eq:mat_a_new} to compute $\mathbf{A}(t)$, the analytic derivation of $\mathbf{A}(t)$ to $t$ is complex enough, not $\rho(\mathbf{A}(t)) \mathbf{Y}(t) + \mathbf{X}(t)\rho(\mathbf{A}(t))$. To address this issue, we utilize a numerical approximation scheme, i.e., computing the derivation of $\mathbf{A}(t)$ to $t$ after the normalization of $\mathbf{x}(t)$ and $\mathbf{y}(t)$. Computation details are shown in Sec. \ref{sec:refine_assign_eq}. We do not integrate $d\rho(\mathbf{A}(t))/dt$, since $\rho(\mathbf{A}(t))$ is a matrix with constraints.

\subsection{ODE-driven based cross-attention}
\label{sec:refine_assign_eq}

To unify Eqs. \ref{eq:refine_prune}, \ref{eq:mat_a_new}, and \ref{eq:ref_propagate_a_mat} into a single framework, we propose the OCA module, as shown in Fig. \ref{fig:solution}. It consists of two stages:

\vspace{+1mm}
\noindent\textbf{Step 1: attention initialization}. Given the image and point cloud features $\mathbf{x}[0]$ and $\mathbf{y}[0]$ (details of them are provided in Sec. \ref{sec:refine_solution}), $\mathbf{A}[0]$ is computed via Eq. \ref{eq:mat_a_new} and then pruned via Eq. \ref{eq:refine_prune}. $\rho_{\text{sparse}}(\mathbf{A}[0])$ is obtained as the initialized attention. 

\vspace{+1mm}
\noindent\textbf{Step 2: attention propagation}. This step is to compute $\mathbf{A}[T]$ through Eq. \ref{eq:ref_propagate_a_mat} discretely. As discussed in Sec. \ref{sec:refine_overview},  the derivation of $\mathbf{A}(t)$ to $t$ is computed after the normalization of $\mathbf{x}(t)$ and $\mathbf{y}(t)$, so Eq. \ref{eq:diff_assign_ode_aux} needs to be corrected as:
\begin{equation}
\label{eq:diff_assign_ode_aux_1}
\mathbf{X}_{\text{norm}}[k]=\mathbf{x}_{\text{norm}}[k]\mathbf{x}_{\text{norm}}[k]^T,\mathbf{Y}_{\text{norm}}[k]=\mathbf{y}_{\text{norm}}[k]\mathbf{y}_{\text{norm}}[k]^T
\end{equation}

A discrete computation approach of Eq. \ref{eq:ref_propagate_a_mat} is provided as: 
\begin{equation}
\label{eq:att_prop_1}
\mathbf{A}[k+1]=\mathbf{A}[k]
+\tau (\rho_{\text{sparse}}(\mathbf{A}[k]) \mathbf{Y}_{\text{norm}}[k] + \mathbf{X}_{\text{norm}}[k] \rho_{\text{sparse}}(\mathbf{A}[k]))
\end{equation}
\begin{equation}
\label{eq:att_prop_2}
\begin{cases}
\mathbf{x}[k+1] = \mathbf{x}[k]+ \tau\rho_{\text{sparse}}(\mathbf{A}[k+1])\mathbf{y}[k] \\
\mathbf{y}[k+1] = \mathbf{y}[k]+ \tau\rho_{\text{sparse}}(\mathbf{A}[k+1]^T)\mathbf{x}[k]
\end{cases}, \,\, k=0,...,T-1
\end{equation}
where $\tau \in [0,1]$ is a time step. During ODE propagation, Eqs. \ref{eq:att_prop_1} and \ref{eq:att_prop_2} are computed iteratively. Finally, the OCA module outputs weighted average features of $\mathbf{x}[0]$ and $\mathbf{x}[T]$, $\mathbf{y}[0]$ and $\mathbf{y}[T]$, with a weighting factor $\omega$. Also, Eqs. \ref{eq:diff_assign_ode_aux_1}, \ref{eq:att_prop_1}, and \ref{eq:att_prop_2} show that OCA is a lightweight and non-parametric module. 

\subsection{OCA-embedded I2P registration}
\label{sec:refine_solution}

To further leverage OCA, we develop a simple yet effective OCA-embedded I2P registration scheme. It is noted that existing I2P registration frameworks consist of three modules, i.e., feature extraction, feature interaction, and feature matching \cite{2d3d-matr, flowi2p, ca-i2p-iccv-25, bridge, ldf-i2p}. To complement existing cross-attention modules, we insert the proposed OCA module after the feature interaction module, as illustrated in the middle image of Fig. \ref{fig:intro}. To supervise the OCA-embedded I2P registration, we adopt a two-stage training strategy. In the first stage, we train the original I2P registration framework from scratch with $\tau=0$. In the next stage, we fine-tune the framework with the original $\tau$. The default I2P registration loss \cite{2d3d-matr} is adopted during the two-stage training. The usage of two-stage training enforces strategy \textbf{(S1)} by ensuring that $\rho_{\text{sparse}}(\mathbf{A}[0])$ is closed to $\mathbf{A}_{gt}$. 



\section{Experiments and Discussions}
\label{sec:exp}

To evaluate the effectiveness of the proposed OCA-embedded I2P registration, we conduct experiments on four public datasets with five state-of-the-art baselines \cite{2d3d-matr, flowi2p, bridge, ca-i2p-iccv-25, ldf-i2p} under the standard, fine-tuning, and zero-shot comparisons. 

\subsection{Configurations}
\label{sec:exp_config}

We illustrate the experiment configurations, including datasets, baselines, metrics, and implementations of I2P registration. 

\vspace{+1mm}
\noindent\textbf{Datasets}. Four public indoor datasets are used for the I2P registration. 7-Scenes \cite{scene-7-dataset} is a common dataset for 2D and 3D registration evaluation. We follow the previous work \cite{2d3d-matr} to decompose the training and testing splits. Then, to verify the stability of different I2P registration models in the new scenes, other datasets, such as RGBD-v2 \cite{RGBD-dataset}, TUM \cite{tum} and ScanNet \cite{scan-net}, are utilized for the fine-tuning comparisons where the ratio of training and testing samples are 1:10. In each dataset, the inputs of I2P registration are RGB images and point clouds (without color features). Input data is visualized and discussed in Appendix C.


\vspace{+1mm}
\noindent\textbf{Baselines}. We conduct comparisons with five state-of-the-art baseline methods, including Matr (ICCV'23) \cite{2d3d-matr}, Flow-I2P (IJCV'25) \cite{flowi2p}, Bridge (AAAI'25) \cite{bridge}, CA-I2P (ICCV'25) \cite{ca-i2p-iccv-25}, and LDF-I2P (TIM'25) \cite{ldf-i2p}. They are feature-based registration methods that contain the various advanced {prior-knowledge-free} cross-attention mechanisms. To evaluate the performance of the proposed method, OCA is embedded into the above I2P registration frameworks as X+OCA. For a fair comparison, we enforce the inputs of all methods to be the same (i.e., RGB images and point clouds) without using any geometric priors. Moreover, to verify the gain stemming from OCA versus more training epochs, we set a method X+\texttt{Ref} which is to train X with the epoch and optimizer settings the same as X+OCA. The comparison between X+\texttt{Ref} and X+OCA is fair, since they do not introduce any new learnable parameters.   

\vspace{+1mm}
\noindent\textbf{Metrics}. To fully measure the I2P registration performance, we use inlier ratio (IR) and registration recall (RR) as the main metrics, since they are common in any registration task. Compared with IR, RR can reflect the global registration accuracy. Moreover, we use the relative rotation error (RRE) and relative translation error (RTE) to evaluate the accuracy of the camera pose estimated from the predicted 2D-3D correspondences. Thresholds of IR and RR are 5cm and 10cm. Computation details of these metrics refer to work \cite{2d3d-matr}. 

\vspace{+1mm}
\noindent\textbf{Implementations}. All methods are trained, fine-tuned, and tested on the same computer with one NVIDIA GeForce RTX 3080 GPU. The learning rate is 1e-4. Training epochs of X and X+OCA (X+\texttt{Ref}) are set as 25 and 32, respectively. Adam optimizer is utilized in the training stage. The maximum number of point clouds is 30K, and RGB images have a size of 320$\times$480. In the comparisons, some methods \cite{2d3d-matr, flowi2p, ldf-i2p} are open-source, while others \cite{bridge, ca-i2p-iccv-25} are implemented by ourselves based on the architecture of \cite{2d3d-matr}. 

\begin{table*}[t]
\centering 
\begin{minipage}{0.45\textwidth}
    \centering
    \caption{Standard evaluation of the proposed OCA module with five baselines on the 7-Scene dataset \cite{scene-7-dataset}.}
    \begin{tabular}{l|cc}
\toprule
Methods & IR & RR \\
\hline
Matr\cite{2d3d-matr} &  0.453 & 0.472 \\
Matr\cite{2d3d-matr}+\texttt{Ref} & 0.475 & 0.501 \\
Matr\cite{2d3d-matr}+OCA & \textbf{0.501} & \textbf{0.552}\textcolor{blue}{$^\mathbf{+5\%}$} \\
\hline
Flow-I2P\cite{flowi2p} &  0.469 & 0.511 \\
Flow-I2P\cite{flowi2p}+\texttt{Ref} & 0.459 & 0.562 \\
Flow-I2P\cite{flowi2p}+OCA &  \textbf{0.530} & \textbf{0.596}\textcolor{blue}{$^\mathbf{+3\%}$} \\
\hline
Bridge\cite{bridge} &  0.460 & 0.505 \\
Bridge\cite{bridge}+\texttt{Ref} & 0.501 & 0.520 \\
Bridge\cite{bridge}+OCA &  \textbf{0.529} & \textbf{0.550}\textcolor{blue}{$^\mathbf{+3\%}$} \\
\hline
CA-I2P\cite{ca-i2p-iccv-25} &  0.459 & 0.495 \\
CA-I2P\cite{ca-i2p-iccv-25}+\texttt{Ref} & 0.503 & 0.534 \\
CA-I2P\cite{ca-i2p-iccv-25}+OCA &  \textbf{0.524} & \textbf{0.565}\textcolor{blue}{$^\mathbf{+3\%}$} \\
\hline
LDF-I2P\cite{ldf-i2p} &  0.446 & 0.541 \\
LDF-I2P\cite{ldf-i2p}+\texttt{Ref} & 0.487 & 0.551 \\
LDF-I2P\cite{ldf-i2p}+OCA &  \textbf{0.525} & \textbf{0.591}\textcolor{blue}{$^\mathbf{+4\%}$} \\
\bottomrule
\end{tabular}
\vspace{-12pt}
    \label{table:exp_1}
\end{minipage}\hfill 
\begin{minipage}{0.45\textwidth}
    \centering
\caption{Fine-tuning evaluation of the proposed OCA module with five baselines on the RGBD-v2 dataset \cite{RGBD-dataset}. }
    \begin{tabular}{l|cc}
\toprule
Methods & IR & RR \\
\hline
Matr\cite{2d3d-matr} &  0.341 & 0.457 \\
Matr\cite{2d3d-matr}+\texttt{Ref} &  0.372 & 0.498 \\
Matr\cite{2d3d-matr}+OCA & \textbf{0.375} & \textbf{0.537}\textcolor{blue}{$^\mathbf{+4\%}$} \\
\hline
Flow-I2P\cite{flowi2p} &  0.361 & 0.547 \\
Flow-I2P\cite{flowi2p}+\texttt{Ref} &  0.379 & 0.573 \\
Flow-I2P\cite{flowi2p}+OCA &  \textbf{0.387} & \textbf{0.580}\textcolor{blue}{$^\mathbf{+1\%}$} \\
\hline
Bridge\cite{bridge} &  0.333 & 0.490 \\
Bridge\cite{bridge}+\texttt{Ref} &  0.350 & 0.510 \\
Bridge\cite{bridge}+OCA &  \textbf{0.354} & \textbf{0.529}\textcolor{blue}{$^\mathbf{+2\%}$} \\
\hline
CA-I2P\cite{ca-i2p-iccv-25} &  0.349 & 0.533 \\
CA-I2P\cite{ca-i2p-iccv-25}+\texttt{Ref} &  0.335 & 0.546\\
CA-I2P\cite{ca-i2p-iccv-25}+OCA &  \textbf{0.362} & \textbf{0.583}\textcolor{blue}{$^\mathbf{+4\%}$} \\
\hline
LDF-I2P\cite{ldf-i2p} &  0.325 & 0.501 \\
LDF-I2P\cite{ldf-i2p}+\texttt{Ref} &  0.340 & 0.527 \\
LDF-I2P\cite{ldf-i2p}+OCA &  \textbf{0.370} & \textbf{0.540}\textcolor{blue}{$^\mathbf{+1\%}$} \\
\bottomrule
\end{tabular}
\vspace{-12pt}
    \label{table:exp_2}
\end{minipage}
\end{table*}

\begin{table*}[t]
\centering 
\begin{minipage}{0.45\textwidth}
    \centering
\caption{Fine-tuning evaluation of the proposed OCA module with five baselines on the TUM dataset \cite{tum}.}
    \begin{tabular}{l|cc}
\toprule
Methods & IR & RR \\
\hline
Matr\cite{2d3d-matr} &  0.568 & 0.472 \\
Matr\cite{2d3d-matr}+\texttt{Ref} & 0.629 & 0.647 \\
Matr\cite{2d3d-matr}+OCA & \textbf{0.703} & \textbf{0.705}\textcolor{blue}{$^\mathbf{+8\%}$} \\
\hline
Flow-I2P\cite{flowi2p} &  0.666 & 0.677 \\
Flow-I2P\cite{flowi2p}+\texttt{Ref} &  0.703 & 0.732 \\
Flow-I2P\cite{flowi2p}+OCA &  \textbf{0.737} & \textbf{0.775}\textcolor{blue}{$^\mathbf{+4\%}$} \\
\hline
Bridge\cite{bridge} &  0.653 & 0.716 \\
Bridge\cite{bridge}+\texttt{Ref} &  0.716 & 0.761 \\
Bridge\cite{bridge}+OCA &  \textbf{0.744} & \textbf{0.806}\textcolor{blue}{$^\mathbf{+4\%}$} \\
\hline
CA-I2P\cite{ca-i2p-iccv-25} &  0.668 & 0.705 \\
CA-I2P\cite{ca-i2p-iccv-25}+\texttt{Ref} &  0.705 & 0.743 \\
CA-I2P\cite{ca-i2p-iccv-25}+OCA &  \textbf{0.744} & \textbf{0.778}\textcolor{blue}{$^\mathbf{+3\%}$}\\
\hline
LDF-I2P\cite{ldf-i2p}&  0.627 & 0.636 \\
LDF-I2P\cite{ldf-i2p}+\texttt{Ref} &  0.643 & 0.675 \\
LDF-I2P\cite{ldf-i2p}+OCA &  \textbf{0.710} & \textbf{0.764}\textcolor{blue}{$^\mathbf{+9\%}$} \\
\bottomrule
\end{tabular}
\vspace{-6pt}
    \label{table:exp_3}
\end{minipage}\hfill 
\begin{minipage}{0.45\textwidth}
    \centering
\caption{Fine-tuning evaluation of the proposed OCA module with five baselines on the ScanNet dataset \cite{scan-net}.}
    \begin{tabular}{l|cc}
\toprule
Methods & IR & RR \\
\hline
Matr\cite{2d3d-matr} &  0.342 & 0.664 \\
Matr\cite{2d3d-matr}+\texttt{Ref} & 0.376 & 0.691 \\
Matr\cite{2d3d-matr}+OCA & \textbf{0.387} & \textbf{0.711}\textcolor{blue}{$^\mathbf{+2\%}$} \\
\hline
Flow-I2P\cite{flowi2p} &  0.386 & 0.763 \\
Flow-I2P\cite{flowi2p}+\texttt{Ref} &  0.391 & 0.770 \\
Flow-I2P\cite{flowi2p}+OCA &  \textbf{0.442} & \textbf{0.789}\textcolor{blue}{$^\mathbf{+2\%}$} \\
\hline
Bridge\cite{bridge} &  0.382 & 0.724 \\
Bridge\cite{bridge}+\texttt{Ref} &  0.385 & 0.733 \\
Bridge\cite{bridge}+OCA &  \textbf{0.395} & \textbf{0.745}\textcolor{blue}{$^\mathbf{+1\%}$} \\
\hline
CA-I2P\cite{ca-i2p-iccv-25} &  0.383 & 0.730 \\
CA-I2P\cite{ca-i2p-iccv-25}+\texttt{Ref} &  0.408 & 0.739 \\
CA-I2P\cite{ca-i2p-iccv-25}+OCA &  \textbf{0.439} & \textbf{0.743}\textcolor{blue}{$^\mathbf{+1\%}$} \\
\hline
LDF-I2P\cite{ldf-i2p} &  0.392 & 0.697 \\
LDF-I2P\cite{ldf-i2p}+\texttt{Ref} &  0.422 & 0.744 \\
LDF-I2P\cite{ldf-i2p}+OCA &  \textbf{0.461} & \textbf{0.783}\textcolor{blue}{$^\mathbf{+4\%}$} \\
\bottomrule
\end{tabular}
\vspace{-6pt}
    \label{table:exp_4}
\end{minipage}
\end{table*}

\begin{table}[!ht]
\centering
\caption{
Zero-shot registration evaluation of the proposed OCA module with representative baselines on the ScanNet dataset \cite{scan-net}. 
}
\resizebox{1\linewidth}{!}{
\begin{tabular}{c|ccc|c|ccc}
\toprule
{Matr} \cite{2d3d-matr} &  +\texttt{Ref} & +OCA & +\texttt{Zero}-OCA & {Flow-I2P} \cite{2d3d-matr} &  +\texttt{Ref} & +OCA & +\texttt{Zero}-OCA \\
\hline
IR & \textbf{0.259} & 0.253 & 0.215 & IR & \textbf{0.310} & 0.287 & 0.210  \\
RR & 0.184 & 0.217\textcolor{blue}{$^\mathbf{+3\%}$} & \textbf{0.276}\textcolor{blue}{$^\mathbf{+9\%}$} & RR & 0.263 & 0.342\textcolor{blue}{$^\mathbf{+8\%}$} & \textbf{0.414}\textcolor{blue}{$^\mathbf{+15\%}$} \\
\hline
{Bridge} \cite{bridge} &  +\texttt{Ref} & +OCA & +\texttt{Zero}-OCA & {LDF-I2P} \cite{ldf-i2p} &  +\texttt{Ref} & +OCA & +\texttt{Zero}-OCA \\
\hline
IR & \textbf{0.330} & 0.307 & 0.295 & IR & \textbf{0.304} & 0.293 & 0.210  \\
RR & 0.224 & \textbf{0.342}\textcolor{blue}{$^\mathbf{+12\%}$} & 0.316\textcolor{blue}{$^\mathbf{+9\%}$} & IR & 0.309 & 0.349\textcolor{blue}{$^\mathbf{+4\%}$} & \textbf{0.428}\textcolor{blue}{$^\mathbf{+12\%}$} \\
\bottomrule
\end{tabular}}
\vspace{-6pt}
\label{table:exp_5}
\end{table}

\subsection{Comparisons}
\label{sec:exp_comp}

\vspace{+1mm}
\noindent\textbf{Standard comparisons}. We first investigate the performance of OCA with five baseline methods on the 7-Scene dataset \cite{scene-7-dataset}. All methods are trained and tested on the training and testing samples split by the work \cite{2d3d-matr}. The means of IR and RR on seven indoor scenes are shown in Table \ref{table:exp_1}. With the help of the OCA module, the IR and RR metrics of baselines are significantly improved. For the Matr \cite{2d3d-matr}, Flow-I2P \cite{flowi2p} and CA-I2P \cite{ca-i2p-iccv-25}, the improvements on RR metric are beyond 3\%. These results demonstrate that the gain of X+OCA stems from OCA rather than the extra training epochs.


\begin{figure}[t]
	\centering
		\includegraphics[width=1.0\linewidth]{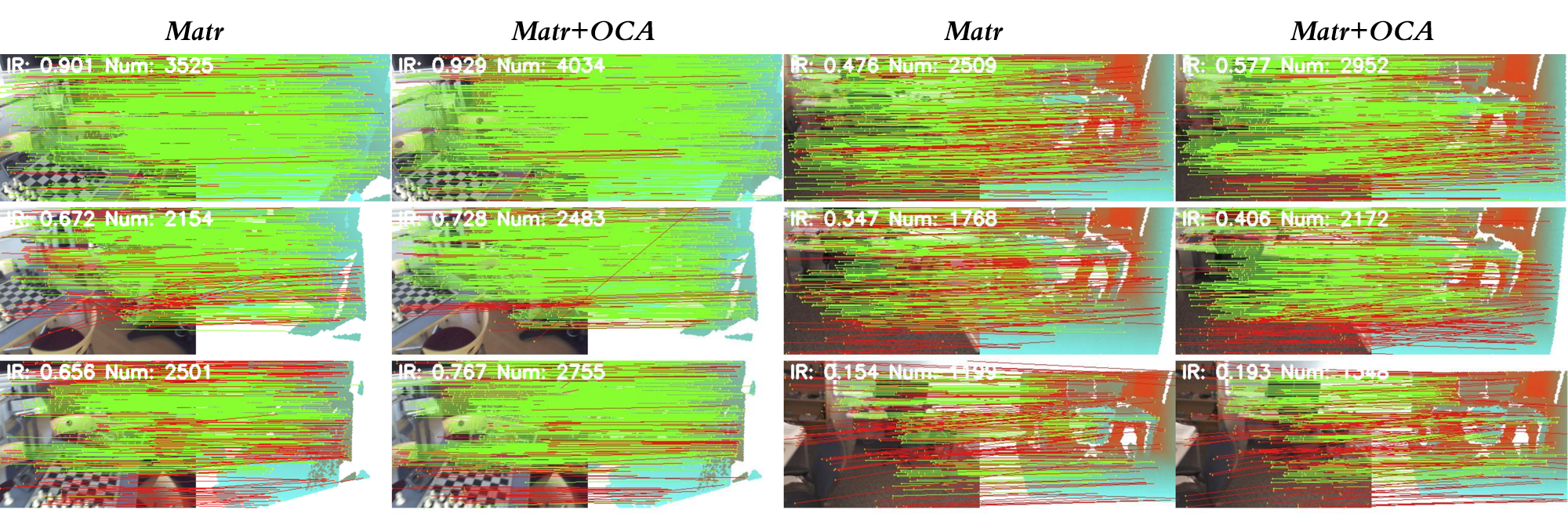}
		\caption{Qualitative comparisons of Matr \cite{2d3d-matr} and Matr+OCA on the 7-Scenes dataset \cite{scene-7-dataset}. Across various scenes with different overlap ratios, the proposed OCA module increases the number of inliers.}
	\label{fig:vis_1}
    \vspace{-12pt}
\end{figure}

\begin{figure}[!ht]
	\centering
		\includegraphics[width=0.92\linewidth]{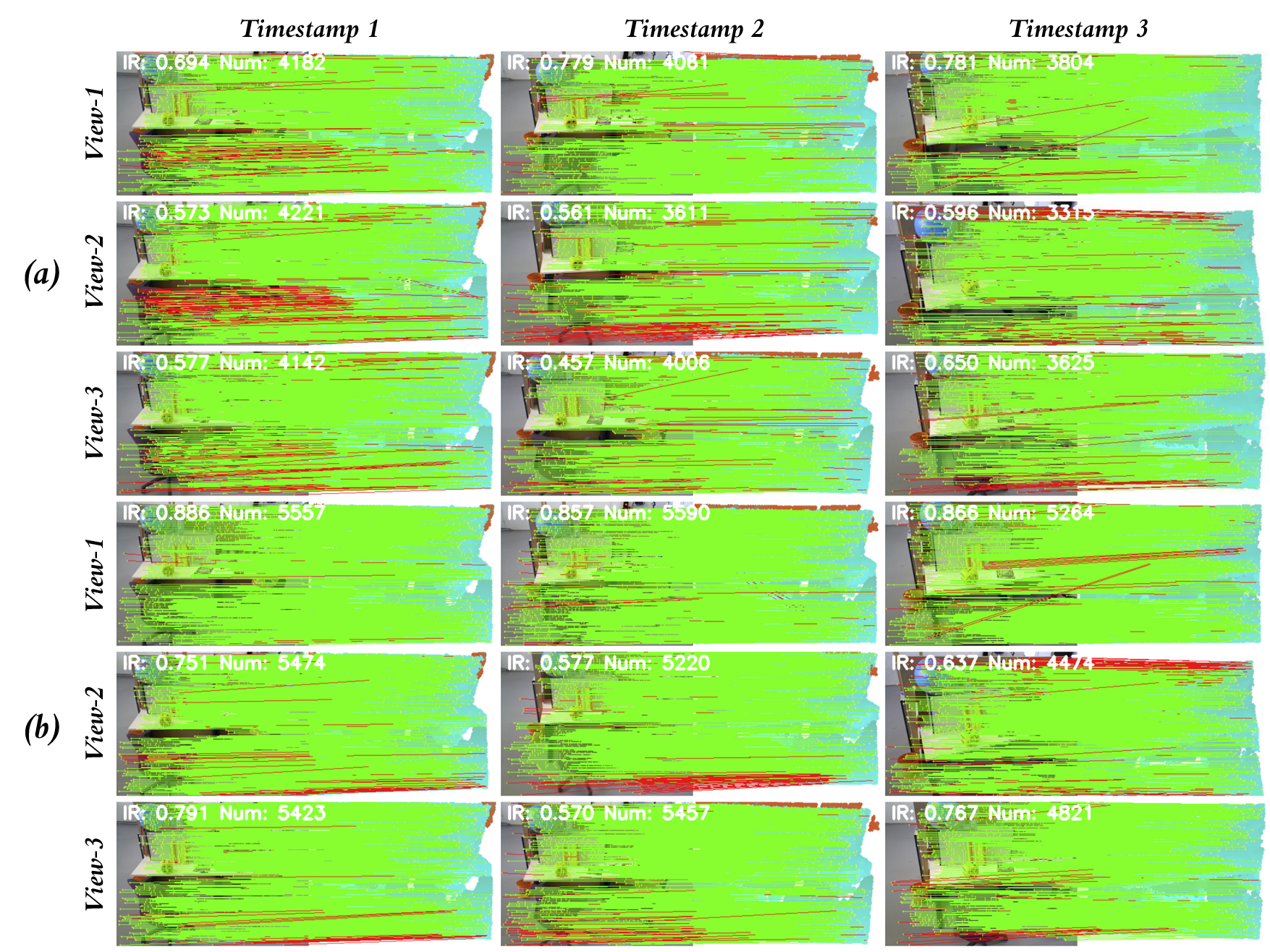}
		\caption{Qualitative comparison on the TUM dataset \cite{tum} under different viewpoints with consecutive timestamps. (a) CA-I2P \cite{ca-i2p-iccv-25} and (b) CA-I2P+OCA. With the integration of OCA, the I2P registration results become more stable and accurate.}
	\label{fig:vis_2}
    \vspace{-12pt}
\end{figure}

\begin{figure}[t]
	\centering
		\includegraphics[width=1\linewidth]{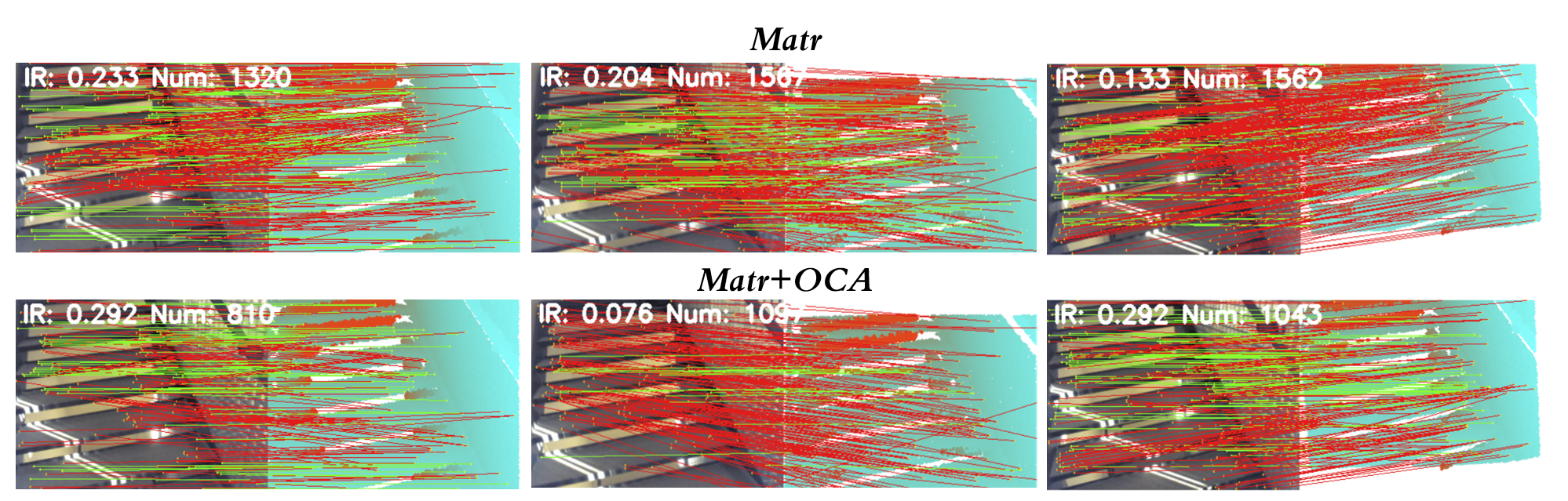}
		\caption{Failure case of the proposed OCA module in extremely texture-less scenes. In the second column, the IR of Matr+OCA decreases to $7.6\%$. }
	\label{fig:vis_app_failure}
    \vspace{-12pt}
\end{figure}

\begin{figure}[t]
	\centering
		\includegraphics[width=1.0\linewidth]{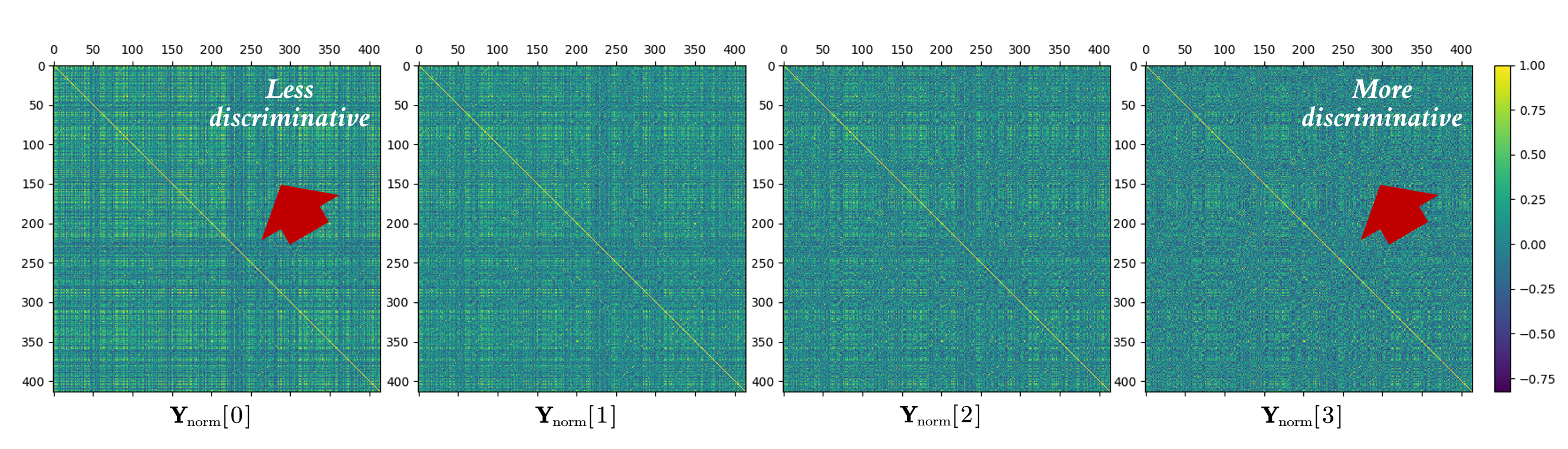}
		\caption{Visualization of $\mathbf{Y}_{\text{norm}}[k]$. As $\mathbf{Y}_{\text{norm}}[k]=\mathbf{y}_{\text{norm}}[k]\cdot\mathbf{y}_{\text{norm}}[k]^T$, a {sparse} $\mathbf{Y}_{\text{norm}}[k]$ indicates that the point feature is less redundant with other points, suggesting greater feature {discriminability}. During ODE propagation, $\mathbf{Y}_{\text{norm}}[k]$ tends to become sparser, which empirically supports the effectiveness of the proposed OCA module.}
	\label{fig:vis_effect}
    \vspace{-6pt}
\end{figure}

\begin{table}[!ht]
\centering
\caption{
Comparisons with existing propagation-based methods on the TUM dataset \cite{tum}. 
}
\begin{tabular}{l|>{\columncolor{blue!16}}cccc}
\toprule
Methods &  Matr+OCA & Matr+\texttt{Simple} DDPM \cite{ddpm} & Matr+\texttt{Simple} FM \cite{flowmatching} & Diff-Reg \cite{diff-reg} \\
\hline
IR & {\textbf{0.703}} & 0.465 & 0.448 & 0.623 \\
RR & {\textbf{0.705}} & 0.460 & 0.459 & 0.602 \\
\bottomrule
\end{tabular}
\vspace{-6pt}
\label{table:exp_6}
\end{table}

\begin{table*}[ht]
\centering 
\begin{minipage}{0.48\textwidth}
    \centering
\caption{Ablation study on the iteration number $T$ on the 7-Scenes dataset \cite{scene-7-dataset}.}
    \begin{tabular}{l|cc>{\columncolor{blue!16}}ccc}
\toprule
$T$ & 1 & 2 & 3 & 4 & 5 \\
\hline
IR & \textbf{0.571} & 0.554 & 0.501 & 0.433 & 0.415 \\
RR & 0.489 & 0.510 & 0.552 & \textbf{0.575} & 0.539 \\
\bottomrule
\end{tabular}
    \label{table:exp_ab_1}
\end{minipage}\hfill 
\begin{minipage}{0.42\textwidth}
    \centering
\caption{Ablation study on the time step $\tau$ on the 7-Scenes dataset \cite{scene-7-dataset}.}
    \begin{tabular}{l|c>{\columncolor{blue!16}}cccc}
\toprule
$\tau$ & 0.05 & 0.10 & 0.15 & 0.20 & 0.25 \\
\hline
IR & \textbf{0.562} & 0.501 & 0.443 & 0.400 & 0.379 \\
RR & 0.495 & \textbf{0.552} & 0.541 & 0.497 & 0.494 \\
\bottomrule
\end{tabular}
    \label{table:exp_ab_2}
\end{minipage}
\end{table*}

\vspace{+1mm}
\noindent\textbf{Fine-tuning comparisons}. To evaluate the performance of the OCA module on I2P registration in the new scenarios, we conduct fine-tuning experiments on multiple datasets. Methods, X, X+OCA and X+\texttt{Ref}, pretrained on the 7-Scenes \cite{scene-7-dataset}, are fine-tuned and tested on the RGBD-v2 \cite{RGBD-dataset}, TUM \cite{tum}, and ScanNet \cite{scan-net} datasets. The comparison results are shown in Tables \ref{table:exp_2}, \ref{table:exp_3}, and \ref{table:exp_4}. As scenes on RGBD-v2 \cite{RGBD-dataset} have less texture and sparser point clouds, the gains of OCA on five baselines are smaller compared with 7-Scenes \cite{scene-7-dataset}. On the TUM  dataset \cite{tum}, the improvements from the OCA module are significant. The RR gain of OCA on Matr \cite{2d3d-matr} exceeds 8\%. Since scenes in the ScanNet dataset \cite{scan-net} contain abundant textures, all compared methods yield promising registration results, with recall rates above 65\%. In particular, X+OCA outperforms all other approaches and obtains state-of-the-art performance. Thus, the fine-tuning results support the effectiveness of the OCA module. 

\vspace{+1mm}
\noindent\textbf{Zero-shot comparisons}. To verify the generalization ability of OCA, we conduct zero-shot I2P registration experiments. Methods, X+\texttt{Ref} and X+OCA, are pretrained on the 7-Scene dataset \cite{scene-7-dataset}, but tested on the ScanNet dataset \cite{scan-net}. We also set a new comparison method X+\texttt{Zero}-OCA. It directly inserts OCA into the X+\texttt{Ref} without any fine-tuning. Results are provided in Table \ref{table:exp_5} with the representative baselines \cite{2d3d-matr, flowi2p, bridge, ldf-i2p}. Both OCA and \texttt{Zero}-OCA significantly improve the registration recall. In the baseline of Flow-I2P \cite{flowi2p}, the gain of \texttt{Zero}-OCA is 15\%. While the gain of OCA on the baseline of Bridge \cite{bridge} is 12\%. It is also found that inlier ratios of OCA and \texttt{Zero}-OCA are smaller than the baseline. The reason is analyzed here. In the {zero-shot} I2P case, initial features are less accurate than in other situations. With inaccurate features, OCA remains high-confidence correspondences while discarding the low-confidence through propagation, and thus IR is dropping. The remaining high-confidence correspondences are accurate and lead to the precise camera pose estimation, and thus RR is finally improved. Overall, these experiments demonstrate the generalization capability of the OCA module.

\begin{table*}[t]
\centering 
\begin{minipage}{0.34\textwidth}
    \centering
\caption{Parameter selection of the $\gamma$.}
    \begin{tabular}{l|c>{\columncolor{blue!16}}ccc}
\toprule
$\gamma$ & 1 & 2 & 4 & 8 \\
\hline
IR & \textbf{0.531} & 0.501 & 0.472 & 0.435 \\
RR & 0.527 & 0.552 & \textbf{0.561} & 0.532 \\
\bottomrule
\end{tabular}
\vspace{-12pt}
    \label{table:exp_ab_3}
\end{minipage}\hfill 
\begin{minipage}{0.34\textwidth}
    \centering
\caption{Parameter selection of the $\omega$.}
    \begin{tabular}{l|c>{\columncolor{blue!16}}ccc}
\toprule
$\omega$ & 0.05 & 0.10 & 0.15 & 0.20  \\
\hline
IR & 0.492 & 0.501 & \textbf{0.527} & 0.511 \\
RR & 0.531 & \textbf{0.552} & 0.542 & 0.525 \\
\bottomrule
\end{tabular}
\vspace{-12pt}
    \label{table:exp_ab_4}
\end{minipage}\hfill
\begin{minipage}{0.24\textwidth}
    \centering
\caption{Runtime analysis (seconds).}
    \begin{tabular}{l|c}
\toprule
 & Time  \\
\hline
Matr\cite{2d3d-matr} & 0.132  \\
Matr+OCA & 0.138  \\
\bottomrule
\end{tabular}
\vspace{-12pt}
    \label{table:exp_ab_5}
\end{minipage}
\end{table*}

\vspace{+1mm}
\noindent\textbf{Qualitative analysis}. Visualizations of some representative baselines Matr \cite{2d3d-matr} and CA-I2P \cite{ca-i2p-iccv-25} w/ and w/o OCA module are presented in Figs. \ref{fig:vis_1} and \ref{fig:vis_2}. More visualizations with the different baselines and datasets are presented in Appendix D. Therefore, the above qualitative results indicate the effectiveness of the OCA module in a variety of scenes. 


\vspace{+1mm}
\noindent\textbf{Comparisons with other propagating methods}. The denoising diffusion probabilistic model (DDPM) \cite{ddpm} and flow matching (FM) \cite{flowmatching} are state-of-the-art methods that can propagate the attention matrix with ODE. As OCA is a non-parametric method, for fair comparisons, we implement the simple versions of DDPM and FM. Diff-Reg \cite{diff-reg}, as a recent work that leverages DDPM for I2P registration, is also compared. Results in Table \ref{table:exp_6} show that the proposed OCA module is more effective than the lightweight DDPM \cite{ddpm} and FM \cite{flowmatching} models as well as Diff-Reg \cite{diff-reg}. In-depth analysis is shown in Appendix E.

\vspace{+1mm}
\noindent\textbf{Extended comparisons}. To further verify the performance of OCA, we conduct two extended comparisons in Appendix F. The first experiment is to compare OCA with existing {prior-knowledge-based} cross-attention-based methods. The second is to test the extensibility of OCA on the 3D registration task. OCA can also strengthen the intra-modal correspondence learning. 

\vspace{+1mm}
\noindent\textbf{Failure case analysis}. One representative failure case is provided in Fig. \ref{fig:vis_app_failure}. In scenes without texture, the initial attention matrix $\rho_{\text{sparse}}(\mathbf{A}[0])$ has larger errors, so any ODE propagation in $\mathbf{A}[0]$ is unstable and inaccurate. In future work, we will design a feedback mechanism to evaluate the accuracy of $\rho_{\text{sparse}}(\mathbf{A}[0])$ and propose a more stable ODE propagation strategy for $\mathbf{A}[T]$. It will enhance the robustness of the ODE-based cross-attention scheme.

\subsection{Ablation studies}
\label{sec:exp_ablation}

Ablations are conducted on the 7-Scenes dataset \cite{scene-7-dataset}. Matr \cite{2d3d-matr} is the baseline.  

\vspace{+1mm}
\noindent\textbf{Attention propagating analysis}. The core of OCA is to propagate the attention matrix $\mathbf{A}(t)$ via  Eq. \ref{eq:ref_propagate_a_mat}. To implement Eq. \ref{eq:ref_propagate_a_mat}, we adopt a discrete propagation scheme to compute $\mathbf{A}[T]$ where the iteration number $T$ and time step $\tau$ are crucial parameters. Ablation of $T$ is provided in Table \ref{table:exp_ab_1}. When $T$ is increasing from $1$ to $4$, it is observed that the inlier ratio drops from 57.1\% to 43.3\% while the registration recall is improved from 48.9\% to 57.5\%. During the propagation of ODE, $\mathbf{A}[T]$ will refine the high-confidence correspondences while discarding the low-confidence correspondences, so that the inlier ratio is decreased, but registration recall is increased. When $T \geq 5$, $\mathbf{A}[T]$ has a risk of overfitting to some 2D-3D correspondences. It causes both drops. So, we set $T$ as $3$ for better balancing both metrics. Ablation of $\tau$ is presented in Table \ref{table:exp_ab_2}. A large $\tau$ causes the instability of the assignment ODE, so we set $\tau$ to $0.10$ for better performance. Besides, the effect of ODE propagation is visualized in Fig. \ref{fig:vis_effect}. The above results support the effectiveness of the OCA module. 

\vspace{+1mm}
\noindent\textbf{Other parameter selection}. Ablations of other parameters $\gamma$ and $\omega$ are shown in Tables \ref{table:exp_ab_3} and \ref{table:exp_ab_4}. The above results validate that a suitable selection of parameters enhances the effectiveness of the OCA module. More in-depth analysis of $\gamma$ and $\omega$ is provided in Appendix G. Table \ref{table:exp_ab_5} shows that the propagation of assignment ODEs only costs 6ms, which verifies the efficiency of the OCA module. 

\section{Conclusions}
\label{sec:conclusions}

In this paper, we address attention ambiguity caused by the cross-modal gap by revisiting cross-attention in I2P registration from the ODE perspective. Specifically, we construct the {assignment ODEs} to approximate ideal cross-modal feature interactions. We establish a conceptual connection between these ODEs and the conventional cross-attention mechanism. By analyzing the stability properties of the assignment ODEs, we develop the ODE-driven cross-attention (OCA) module, which enhances correspondence learning through iterative refinement within existing I2P registration frameworks. Finally, extensive experiments on four public datasets with five representative baselines demonstrate the effectiveness and generalizability of the proposed OCA module. 

\vspace{+1mm}
\noindent\textbf{Limitation and future work}. Although OCA is {robust} to initial features with the {moderate noises} (cross-dataset case in Tables \ref{table:exp_3} and \ref{table:exp_4}) and {large noises} (zero-shot case in Table \ref{table:exp_5}), it fails in the heavily noised cases as shown in Fig. \ref{fig:vis_app_failure}. This limitation arises from the reliance of ODE propagation on a reasonably informative initialization. In future work, we plan to extend the theoretical framework and develop a more robust ODE-based feature interaction mechanism to further improve reliability in challenging I2P registration scenarios.

\vspace{+1mm}
\noindent\textbf{Acknowledgments}. This work is partially supported by the National Natural Science Foundation of China (Grand ID: 62502171, 62372377), General Program of the Natural Science Basic Research Plan of Shaanxi Province (Grand ID: 2025JC-YBMS-651), and China Postdoctoral Science Foundation (Grand ID: 2024M761014, GZC20252285)

\appendix
\subsection*{A. Derivation of Eq. 5 and discussion of convergence condition (c2)}

\noindent\textbf{Derivation of Eq. 5}. The core of Eq. 5 is to derive the time derivative $d\mathbf{A}(t)/dt$. Given $\mathbf{A}(t)=\mathbf{x}(t)\mathbf{y}(t)^T$, we compute $d\mathbf{A}(t)/dt$ using the product rule for matrix differentiation as follows:
\begin{equation}
\label{eq:a1}
\frac{d\mathbf{A}(t)}{dt} = \frac{d\mathbf{x}(t)\mathbf{y}(t)^T}{dt}=\left(\frac{d\mathbf{x}(t)}{dt}\right)\mathbf{y}(t)^T+\mathbf{x}(t)\left(\frac{d\mathbf{y}(t)}{dt}\right)^T
\end{equation}
Substituting the definitions of $d\mathbf{x}(t)/dt$ and $d\mathbf{y}(t)/dt$ into Eq. \ref{eq:a1}, we obtain:
\begin{equation}
\label{eq:a2}
\frac{d\mathbf{A}(t)}{dt} = \rho(\mathbf{A}(t)) \mathbf{y}(t) \mathbf{y}(t)^T + \mathbf{x}(t) \mathbf{x}(t)^T \rho(\mathbf{A}(t)) 
= \rho(\mathbf{A}(t)) \mathbf{Y}(t) + \mathbf{X}(t) \rho(\mathbf{A}(t)) 
\end{equation}

Finally, Eq. 5 (in the main paper) is directly derived from Eq. \ref{eq:a2}. 

\vspace{+1mm}
\noindent\textbf{Discussion of condition (c2)}. We explain why \textbf{condition (C2)} does not significantly contribute to enhancing I2P registration performance. According to the theory of the Sylvester equation \cite{Sylvester}, equation $\rho(\mathbf{A}(t)) \mathbf{Y}(t) + \mathbf{X}(t) \rho(\mathbf{A}(t))=\mathbf{0}$ has a solution only if $\mathbf{X}(t)$ and $\mathbf{Y}(t)$ share at least one common eigenvalue. However, in the practical learning of deep neural networks (DNNs), it is difficult to ensure that feature matrices $\mathbf{X}(t)$ and $\mathbf{Y}(t)$ share at least one common eigenvalue. Even if the condition is satisfied, the solution to $\rho(\mathbf{A}(t)) \mathbf{Y}(t) + \mathbf{X}(t) \rho(\mathbf{A}(t))=\mathbf{0}$ has a weak correlation with the ground truth assignment matrix $\mathbf{A}_{gt}$. Therefore, based on the above discussion, \textbf{condition (C2)} contributes very little to the design of ODE-driven cross-attention.




\subsection*{B. Explanation of existing solutions from an ODE perspective}

Recent works leverage priors predicted by visual foundation models to enhance cross-attention in I2P registration. We explain their effectiveness from the proposed ODE-based framework.

\begin{figure}[t]
	\centering
		\includegraphics[width=1.0\linewidth]{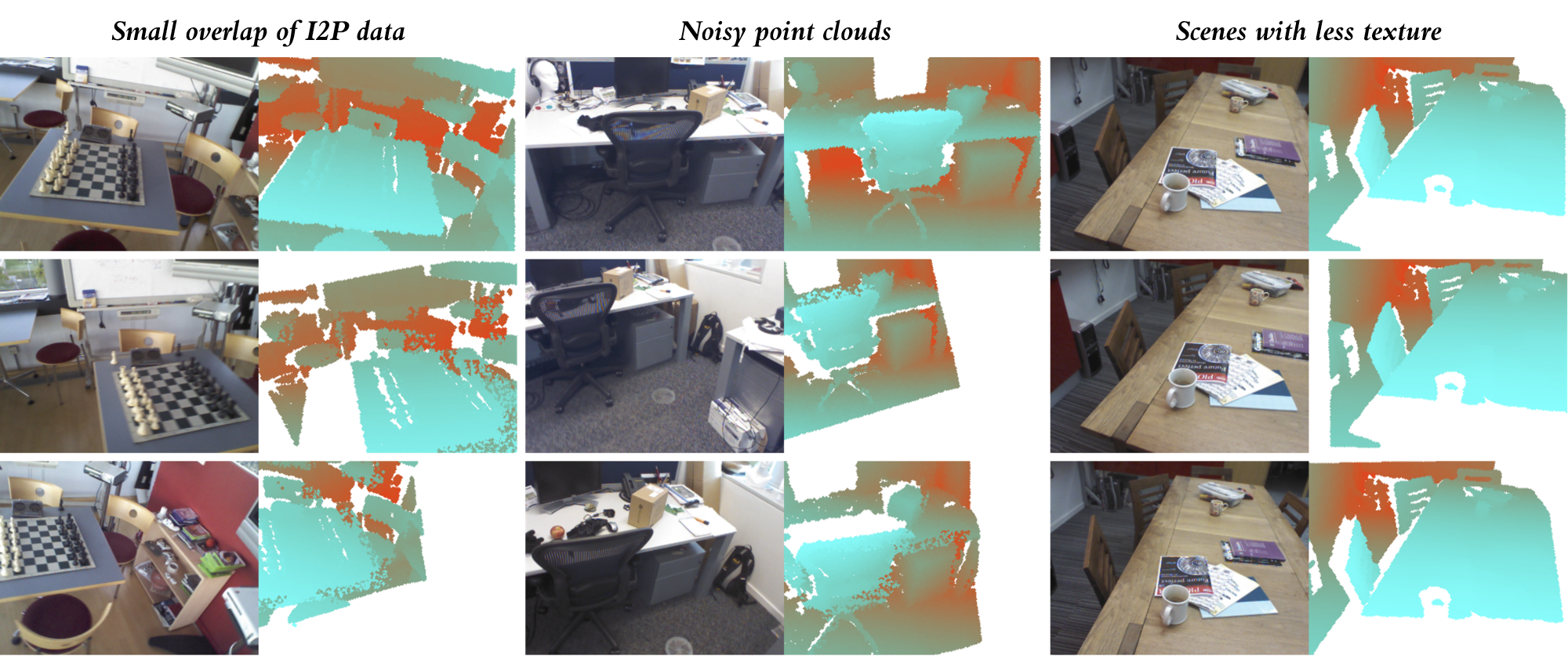}
		\caption{Visualization of input data (RGB image and 3D point cloud) across different scenes. Three key challenges for real-world I2P registration are: (1) small overlap between the image and point cloud, (2) noisy point clouds, and (3) low-texture scenes. }
	\label{fig:vis_app_1}
\end{figure}

\begin{figure}[!ht]
	\centering
		\includegraphics[width=1\linewidth]{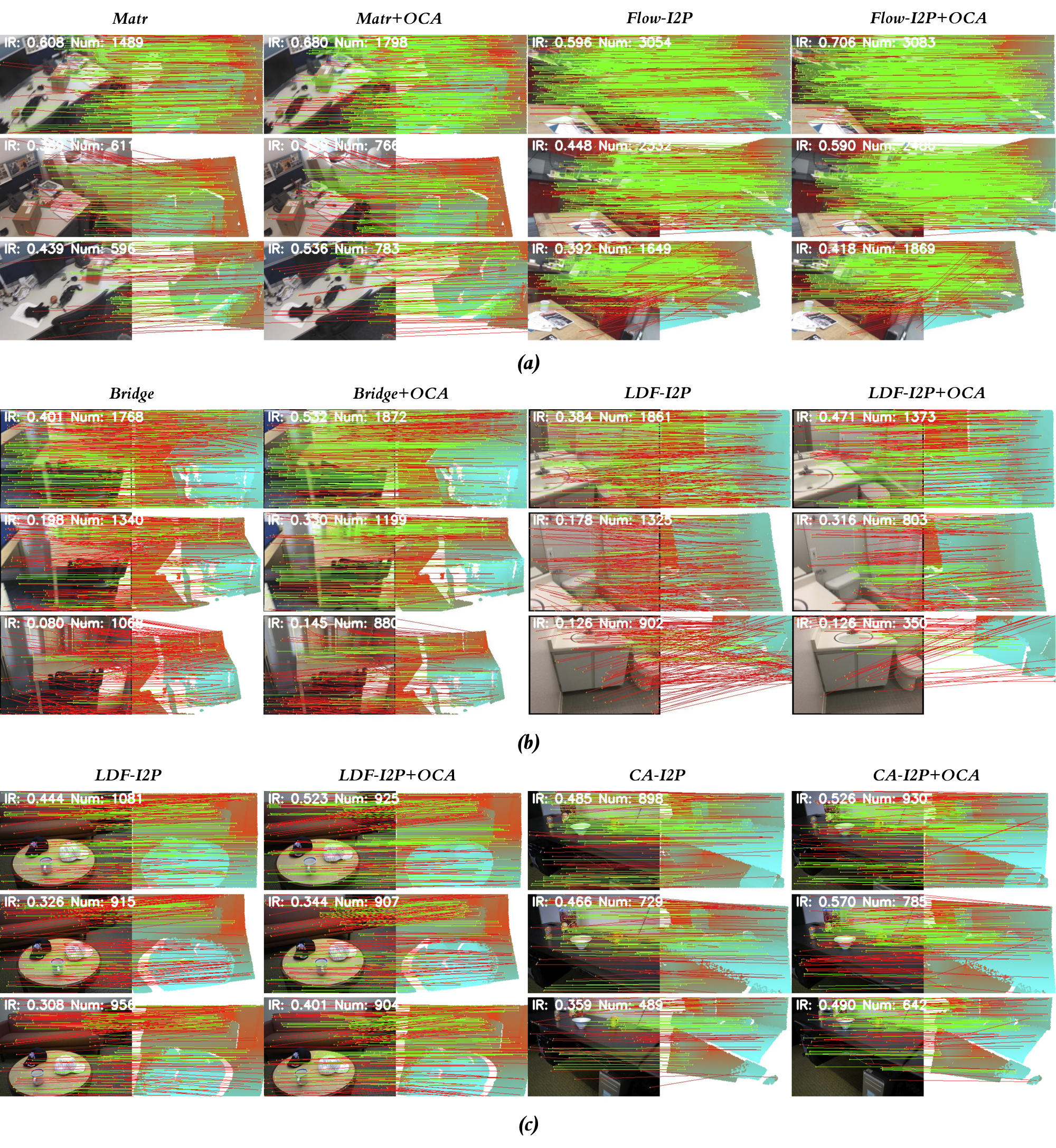}
		\caption{Additional qualitative comparisons of the proposed OCA module integrated with five representative baselines: Matr~\cite {2d3d-matr}, Flow-I2P~\cite {flowi2p}, Bridge~\cite {bridge}, LDF-I2P~\cite {ldf-i2p}, and CA-I2P~\cite {ca-i2p-iccv-25}. Results are shown for (a) 7-Scenes~\cite {scene-7-dataset}, (b) ScanNet~\cite {scan-net}, and (c) RGBD-v2~\cite {RGBD-dataset} datasets. Our ODE-based cross-attention module significantly increases the number of inliers across diverse scenarios.}
	\label{fig:vis_app_2}
\end{figure}

\vspace{+1mm}
\noindent\textbf{Geometric priors}: Common geometric priors include  monocular depths \cite{dep_anything_v2} and surface normals \cite{normal_estimation}. Depth priors enable 3D scene reconstruction from a single RGB image; since the reconstructed 3D scene and the input point cloud approximately follow a 3D similarity transformation~\cite {diff-reg}, depth features reduce the cross-modal gap in I2P data, allowing predicted depth to satisfy the objective of strategy \textbf {(S2)}. Similarly, surface normals predicted from images and computed from point clouds approximately follow a 3D rotation transformation~\cite {flowi2p}; surface normal features thus also mitigate the cross-modal gap, facilitating the achievement of strategy \textbf {(S2)}'s objective.

\vspace{+1mm}
\noindent\textbf{Semantic priors}: This type of prior refers to the use of 2D and 3D semantic (or instance) segmentation models to segment the image and point cloud as several 2D and 3D regions with various semantic labels \cite{sem-align}. Valid 2D-3D correspondences are only established between 2D and 3D regions with identical semantic labels. Semantic priors therefore enforce sparsity in $\rho(\mathbf{A}(t))$, thereby achieving the objective of strategy \textbf{(S1)}.

\vspace{+1mm}
\noindent\textbf{Topological priors}: Topological priors characterize the overlapped relation between 2D and 3D regions \cite{deep-i2p, corr-tcsvt}. In practice, 2D and 3D regions can be segmented by 2D and 3D Segmentation Anything Model (SAM) \cite{top-i2p}. Compared to semantic priors~\cite {sem-align}, topological priors exhibit stronger generalization capability. Valid 2D-3D correspondences are only established between overlapping 2D and 3D regions, so topological priors also enforce sparsity in $\rho(\mathbf{A}(t))$ \cite{top-i2p}, achieving the objective of strategy \textbf {(S1)}.

Theoretically, a method incorporating all the aforementioned priors would achieve optimal I2P registration performance. However, in the practical applications, all these priors are subject to prediction errors (e.g., inaccurate depth estimates, erroneous normals, incorrect semantic labels, and improperly segmented regions), which limit the effectiveness of cross-modal feature interaction for I2P registration. Motivated by these limitations, we propose a prior-knowledge-free module to enhance cross-attention for I2P registration.

\begin{figure}[t]
	\centering
		\includegraphics[width=1.0\linewidth]{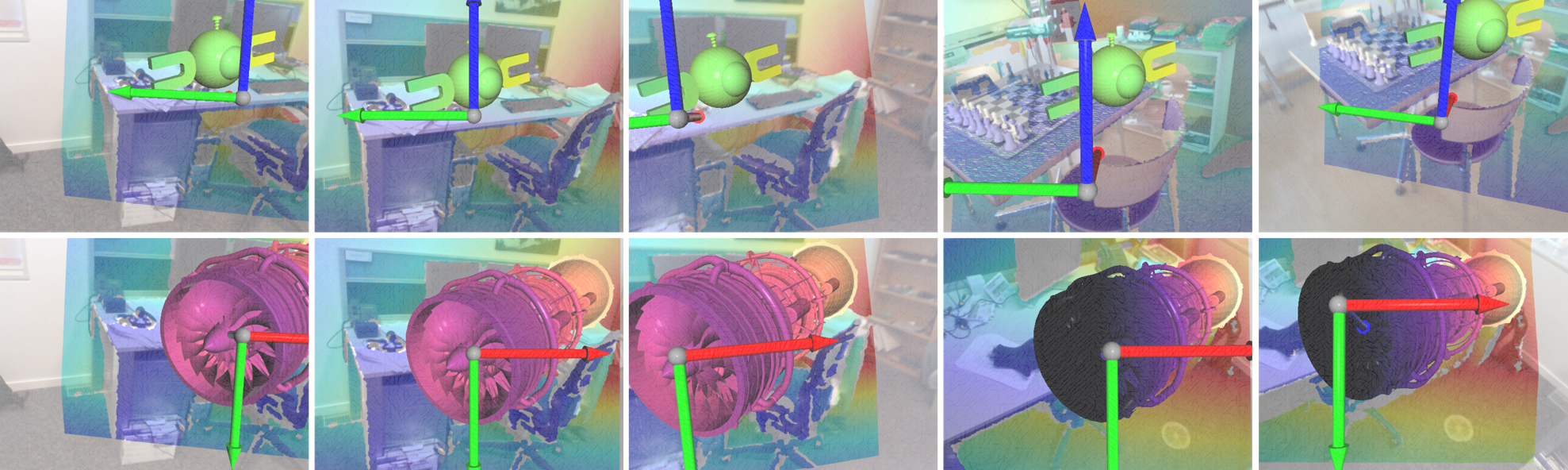}
		\caption{I2P registration for augmented reality (AR) applications. With the OCA layer, virtual objects are rendered with high spatiotemporal consistency with the real world across continuous timestamps.}
	\label{fig:app}
    \vspace{-6pt}
\end{figure}

\subsection*{C. Visualization of input data}

Input data, such as the RGB image and the 3D point cloud, is visualized in Fig. \ref{fig:vis_app_1}. Since RGB images and 3D point clouds are captured in real-world environments, I2P registration faces three challenges: (i) {small overlap} between images and point clouds, (ii) {noise} in 3D point clouds introduced by range sensors, and (iii) {low texture} in certain indoor scenes. In the experiments of the main paper, we validate the effectiveness of OCA for I2P registration methods under these challenging conditions.

\subsection*{D. More qualitative comparisons}

We perform additional qualitative evaluations on the 7-Scenes~\cite {scene-7-dataset}, ScanNet~\cite {scan-net}, and RGBD-v2~\cite {RGBD-dataset} datasets using five representative baseline methods: Matr~\cite {2d3d-matr}, Flow-I2P~\cite {flowi2p}, Bridge~\cite {bridge}, LDF-I2P~\cite {ldf-i2p}, and CA-I2P~\cite {ca-i2p-iccv-25}. Results on challenging scenes are shown in Fig. \ref{fig:vis_app_2}. Our results demonstrate that the proposed OCA module suppresses outliers and increases the number of inliers via attention matrix propagation, while achieving robust performance across diverse complex scenarios. Furthermore, we showcase an application in augmented reality (AR) (see Fig.~\ref {fig:app}): OCA improves registration recall, ensures accurate camera pose estimation, and thus enables high-quality virtual object rendering.

\subsection*{E. In-depth discussion of other propagating methods}

While our method is the first to investigate cross-attention from an ODE perspective, other propagation-based frameworks, including the well-known denoising diffusion probabilistic models (DDPM)~\cite{ddpm} and flow matching (FM)~\cite{flowmatching}, can be adapted to propagate the attention matrix  $\mathbf{A}[k]$. Motivated by this insight, we conducted the experiments reported in Table 6 of the main paper. Due to space constraints in the main paper (only 14 pages), we provide a detailed analysis of these results here. As shown in Table 6, both DDPM~\cite {ddpm} and FM~\cite {flowmatching} degrade I2P registration performance. This is because the original DDPM~\cite {ddpm} and FM~\cite {flowmatching} lack an effective mechanism to enforce sparsity and accuracy of $\mathbf{A}[k]$, making it difficult to enhance the discriminability of 2D and 3D features.

We further compare Matr+OCA with Diff-Reg~\cite {diff-reg}. In the original Diff-Reg framework, accurate metric depth is required to denoise  $\mathbf{A}[k]$ during backward diffusion. Specifically, accurate metric depth enables lifting 2D-3D correspondences to 3D-3D correspondences, allowing outliers to be identified via the iterative closest point (ICP) algorithm~\cite {diff-reg}. However, accurate metric depth constitutes a strong geometric prior. To ensure fair comparison, we modified Diff-Reg~\cite {diff-reg} to use only a U-Net-based neural network to denoise 
$\mathbf{A}[k]$ from $\mathbf{A}[k+1]$ during backward diffusion. Without metric depth, Diff-Reg~\cite {diff-reg} also lacks a mechanism to enforce sparsity and accuracy of $\mathbf{A}[k]$, resulting in performance inferior to Matr+OCA.

\begin{table}[t]
\centering
\caption{
Comparisons with the state-of-the-art prior-knowledge-based registration method \cite{top-i2p} on the TUM dataset \cite{tum}. 
}
\begin{tabular}{c|cccc}
\toprule
Methods &  Matr \cite{2d3d-matr} & Matr \cite{2d3d-matr}+OCA & Top-I2P \cite{top-i2p} & Top-I2P \cite{top-i2p}+OCA \\
\hline
IR & 0.568 & {{0.703}} & 0.683 & \textbf{0.742} \\
RR & 0.472 & {{0.705}} & 0.732 & \textbf{0.764} \\
\bottomrule
\end{tabular}
\vspace{-6pt}
\label{table:exp_extend_1}
\end{table}

\begin{table}[t]
\caption{Verification of the proposed OCA layer in 3D point cloud registration task on the 3DMatch dataset \cite{3Dmatch-dataset}.}
\centering
\begin{tabular}{c|ccccc}
\toprule
\textbf{Number of Sampled Points} & 5000 & 2500 & 1000 & 500 & 250 \\
\hline
\multicolumn{6}{c}{{Inlier Ratio $\uparrow$}} \\
\hline
GeoTransformer \cite{geo-tf} & 71.9\% & 75.2\% & 76.0\% & 82.2\% & 85.1\%  \\
\textbf{GeoTransformer+OCA (Ours)} & \textbf{73.2}\% & \textbf{75.8}\% & \textbf{77.1}\% & \textbf{82.7}\% & \textbf{87.3}\%  \\
\hline
\multicolumn{6}{c}{{Registration Recall $\uparrow$}} \\
\hline
GeoTransformer \cite{geo-tf} & 92.0\% & 91.8\% & 91.8\% & 91.4\% & 91.2\%  \\
\textbf{GeoTransformer+OCA (Ours)} & \textbf{92.4}\% & \textbf{92.3}\% & \textbf{92.6}\% & \textbf{91.7}\% & \textbf{92.2}\%  \\
\bottomrule
\end{tabular}
\label{tab:ape_1}
\vspace{-6pt}
\end{table}

\subsection*{F. Extended comparisons}

\vspace{+1mm}
\noindent\textbf{Comparisons with prior knowledge}. We compare our proposed prior-knowledge-free OCA module with the state-of-the-art (SOTA) prior-knowledge-based method Top-I2P~\cite {top-i2p} on the TUM dataset~\cite {tum}. Top-I2P employs 2D and 3D Segment Anything Models (SAMs) to partition images and point clouds into 2D and 3D regions. It then predicts topological relationships between these 2D and 3D regions and leverages these relationships to facilitate I2P feature matching~\cite {top-i2p}. Results are presented in Table~\ref {table:exp_extend_1}. With the integration of OCA, even the classical baseline Matr~\cite {2d3d-matr} achieves a Registration Recall (RR) metric close to that of Top-I2P~\cite {top-i2p}. Furthermore, when equipped with the OCA module, both the Inlier Ratio (IR) and RR metrics of Top-I2P~\cite {top-i2p} are further improved. These results demonstrate that our proposed prior-knowledge-free OCA module can achieve performance nearly comparable to that of prior-knowledge-based methods.


\vspace{+1mm}
\noindent\textbf{OCA on 3D registration}. In the main paper, OCA is proposed to mitigate attention ambiguity arising from cross-modal discrepancies. Notably, OCA can be seamlessly integrated into existing registration architectures for 3D point cloud registration tasks. To validate this extended capability, we use GeoTransformer~\cite {geo-tf} as the baseline method. For this experiment, we first pre-trained GeoTransformer and then fine-tuned the GeoTransformer+OCA model using the same experimental configuration described in Sec. 5.1 of the main paper. Experimental results on the 3DMatch dataset~\cite {3Dmatch-dataset} are presented in Table~\ref {tab:ape_1}, where both the Inlier Ratio (IR) and Registration Recall (RR) metrics are improved with the integration of the OCA module. This result suggests that OCA is not limited to cross-modal matching but can also enhance intra-modal geometric correspondence estimation. This demonstrates OCA's strong generalization across different registration tasks.

\subsection*{G. More analysis of parameters $\gamma$ and $\omega$}

Due to space constraints in the main paper, we conduct the in-depth analysis of $\gamma$ and $\omega$ corresponding to Tables 9 and 10. $\gamma$ in Eq. 7 regulates the sparsity of $\rho_{\text{sparse}}(\mathbf{A}[k])$. As $\gamma$  increases from  $1$ to $4$, $\rho_{\text{sparse}}(\mathbf{A}[k])$ becomes progressively sparser, focusing attention on high-confidence 2D-3D correspondences while disregarding low-confidence ones. This results in a decrease in Inlier Ratio (IR) but an improvement in Registration Recall (RR). However, when $\gamma$ exceeds $4$, $\rho_{\text{sparse}}(\mathbf{A}[k])$ becomes excessively sparse, making it highly sensitive to the accuracy of the initial matching matrix  $\mathbf{A}[0]$. Consequently, both IR and RR decline for $\gamma > 4$. To balance IR and RR effectively, we set $\gamma=2$ in all experiments. 

We next analyze parameter $\omega$, which fuses 2D and 3D features as $(1-\omega)\mathbf{X}[0]+\omega\mathbf{X}[T]$ and $(1-\omega)\mathbf{Y}[0]+\omega\mathbf{Y}[T]$. As shown in Table 10, performance degrades for $\omega \geq 0.20$, indicating that $\mathbf{X}[T]$ and $\mathbf{Y}[T]$ must be fused carefully to achieve optimal and stable performance. While this behavior is non-intuitive, we explain it from a deep learning perspective. ODE propagation impairs the effectiveness of loss backpropagation. We illustrate this using 2D features as an example:
\begin{equation}
\mathbf{X}_{new}=(1-\omega)\mathbf{X}[0]+\omega\mathbf{X}[T]
\end{equation}

Let  $\mathbf{\Theta}$ denote the set of learnable parameters. The partial derivative of $\mathbf{X}_{new}$ with respect to $\mathbf{\Theta}$ is:
\begin{equation}
\begin{aligned}
\frac{\partial \mathbf{X}_{new}}{\partial \mathbf{\Theta}}&=(1-\omega)\frac{\partial\mathbf{X}[0]}{\partial\mathbf{\Theta}}+\omega\frac{\partial\mathbf{X}[T]}{\partial\mathbf{\Theta}}\\
&= (1-\omega)\frac{\partial\mathbf{X}[0]}{\partial\mathbf{\Theta}}+ \omega \frac{\partial\mathbf{X}[0]}{\partial\mathbf{\Theta}} \cdot \frac{\partial\mathbf{X}[1]}{\partial\mathbf{X}[0]} \cdot \dots \cdot \frac{\partial\mathbf{X}[T]}{\partial\mathbf{X}[T-1]} \\
&= (1-\omega)\frac{\partial\mathbf{X}[0]}{\partial\mathbf{\Theta}}+ \omega \frac{\partial\mathbf{X}[0]}{\partial\mathbf{\Theta}} \cdot \prod_{k=1}^T \frac{\partial\mathbf{X}[k]}{\partial\mathbf{X}[k-1]} \\
\end{aligned}
\end{equation}
During ODE propagation, $\mathbf{X}[k]$ and $\mathbf{Y}[k]$  are tightly coupled, causing the stacked derivative term $\prod_{k=1}^T \frac{\partial\mathbf{X}[k]}{\partial\mathbf{X}[k-1]}$ to easily induce gradient instability. This analysis justifies the selection of a suitable $\omega$ in practical applications. 


%
%
\bibliographystyle{splncs04}
\bibliography{ref}
\end{document}